\documentclass[10pt,twocolumn,letterpaper]{article}

\usepackage[pagenumbers]{cvpr}

\usepackage[T1]{fontenc} 
\usepackage{fontawesome5} 
\usepackage[section]{placeins} 
\AtBeginDocument{\hfuzz=0pt\vfuzz=0pt}
\newcommand{\method}{\textnormal{\textsc{vMFProto}}\xspace}

\newcommand{\papertitle}{Beyond Points: Spherical Distributional Part Prototypes for Interpretable Classification}
\newcommand{\paperauthornames}{Duarte Leão, Diogo Pereira Araújo, Catarina Barata, Carlos Santiago}
\newcommand{\paperauthors}{%
  Duarte Leão\textsuperscript{1,2}\quad
  Diogo Pereira Araújo\textsuperscript{1,2}\quad
  Catarina Barata\textsuperscript{1}\quad
  Carlos Santiago\textsuperscript{1}\\[0.3em]
  \textsuperscript{1}Instituto Superior Técnico\quad
  \textsuperscript{2}Carnegie Mellon University\\[0.35em]
  {\small\href{https://duarte-leao.github.io/vMF-Proto_page/}{%
    \faGlobe\enspace Website}\quad
    \href{https://github.com/Duarte-leao/vMF-Proto}{%
      \faGithub\enspace Code}}
}

\definecolor{cvprblue}{rgb}{0.21,0.49,0.74}
\usepackage[pagebackref,breaklinks,colorlinks,allcolors=cvprblue]{hyperref}
\hypersetup{pdfauthor={\paperauthornames},pdftitle={\papertitle}}

\title{\papertitle}
\author{\paperauthors}

\begin{document}
\maketitle

\begin{abstract}
Prototype-based neural networks aim to make predictions intrinsically explainable by grounding them in a small set of part prototypes. Most methods, however, represent each prototype as a single point in feature space, giving the model a typical appearance to match without describing how much the part varies. Covering that variation with several points can lead multiple prototypes to highlight the same region, making them redundant. We therefore propose \method{}, which uses a von Mises--Fisher distribution on the hypersphere to capture that variation within a single prototype. To learn these prototypes, we use entropic optimal transport to guide discovery and then fit each distribution to its assigned features. The fitted concentration describes how much those features vary, while a separate weight determines how much a match contributes to a class prediction. This separation lets us extend the class-specific model to a shared pool in which classes draw on the same fitted prototypes but learn their own weights. Experiments with frozen DINO backbones on CUB-200-2011 and AP-10K show that both the class-specific and shared-pool models provide stronger part explanations overall, with less redundant evidence. Classification accuracy remains competitive on both datasets as well as on Stanford Dogs and Stanford Cars.
\end{abstract}

\section{Introduction}
\label{sec:intro}

\begin{figure}[tb]
  \centering
  \begin{subfigure}[t]{0.49\linewidth}
    \vspace{0pt}
    \centering
    \includegraphics[width=\linewidth]{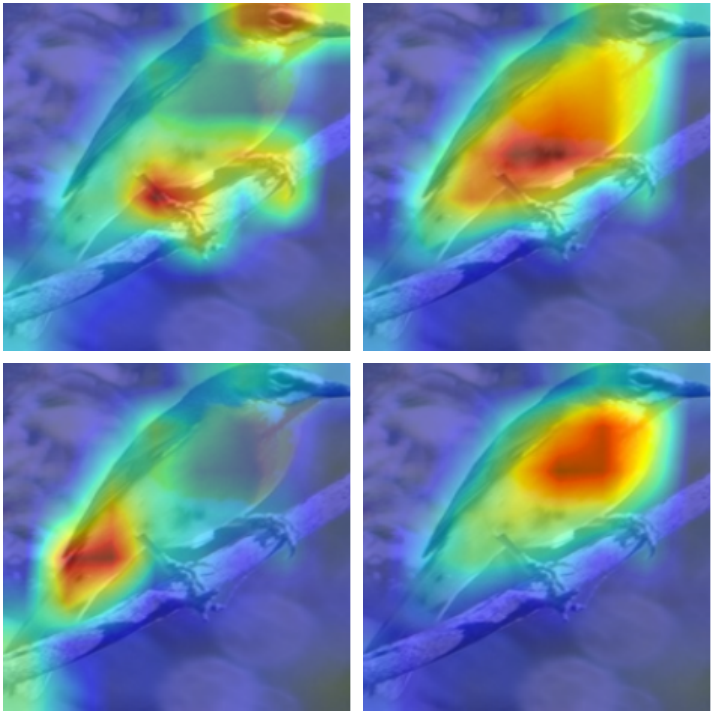}
    \caption{\textbf{ProtoPNet-style baseline.}}
    \label{fig:fig1-a}
  \end{subfigure}
  \hfill
  \begin{subfigure}[t]{0.49\linewidth}
    \vspace{0pt}
    \centering
    \includegraphics[width=\linewidth]{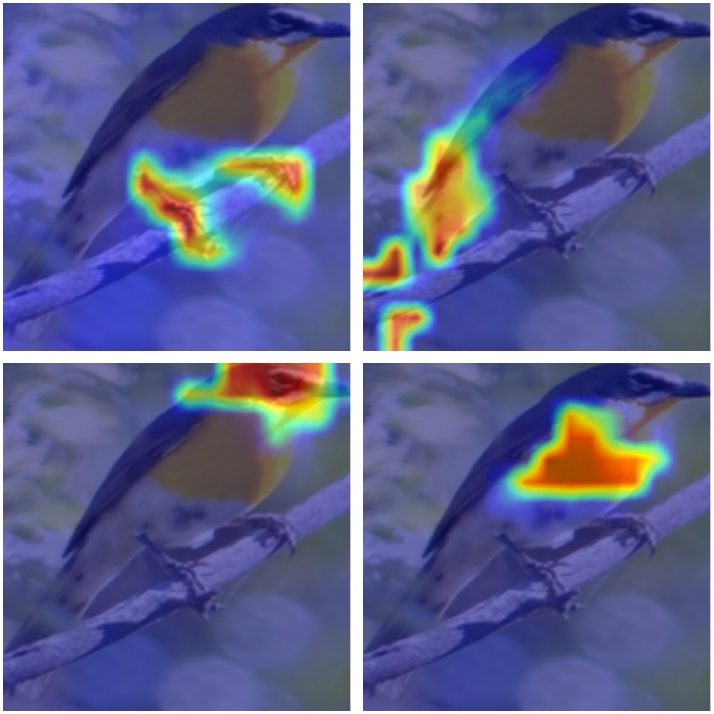}
    \caption{\textbf{Ours.}}
    \label{fig:fig1-b}
  \end{subfigure}

  \caption{\textbf{Why distributional prototypes?} Top four activation maps per model. The point-prototype baseline~\cite{zhu2025nonparampartprototypes} (a) has more overlap than \method{} (b).}
  \label{fig:fig1}
\end{figure}

In fine-grained image classification, a class label tells us what a model predicts but not which parts of the image support that decision. Prototype-based classifiers make those parts available for inspection by comparing local image features with learned references, called prototypes~\cite{chen2019thislookslikethat,nauta2023pipnet,rymarczyk2022protopool}. If a patch on a bird's wing matches a prototype, the explanation can show both where the match occurred and which familiar pattern the model found there. Because that match contributes to the class score, the explanation lets us inspect evidence used in the prediction itself rather than highlight influential regions post hoc~\cite{selvaraju2017gradcam}.

For a wing prototype to provide that evidence across images, it needs to find the wing even when its appearance changes. Pose, lighting, and deformation can all change what a part looks like, but different parts vary in different ways and to different degrees. A point prototype, however, gives the model just one representative appearance to match and cannot describe the variation inherent in that part. Adding more point prototypes can cover appearances that one prototype misses, but several of them may then respond to the same part and repeat the evidence rather than add to it. Evaluations of part-prototype models have documented these redundant responses as well as matches to background cues instead of the intended parts~\cite{huang2023evalprotopnet}.

Instead of adding points to cover that variation, we let each prototype describe a range of appearances. In \method{}, we represent that range with a distribution whose spread can differ from one prototype to the next, unlike in a Gaussian mixture with a shared covariance~\cite{wang2025mgproto}. The choice of distribution follows from how we compare patches. Features from vision foundation models such as DINO and CLIP~\cite{caron2021dino,oquab2024dinov2,simeoni2025dinov3,radford2021clip} are often compared by cosine similarity, which depends on their directions rather than their magnitudes. We therefore normalize patch features to lie on the unit sphere and model them with a von Mises--Fisher (vMF) distribution~\cite{banerjee2005vmf}. Its mean direction $\boldsymbol{\mu}$ describes a typical feature, and its concentration $\kappa$ controls the spread around that direction. The score for a patch then depends on both its angle to the mean and the concentration of the prototype. \Cref{fig:fig1} illustrates what this means for the explanation. On the same image, the distributional prototypes respond to more distinct regions than the point-prototype baseline, providing less redundant evidence.

To learn a distribution for each prototype, we first identify the patches it should describe. A label-free mask restricts the candidates to the foreground, and entropic optimal transport (OT)~\cite{cuturi2013sinkhorn,peyre2019computationalot} guides their assignment to prototypes during discovery, following prior work~\cite{zhu2025nonparampartprototypes}. Once each prototype has a group of patches, we estimate its mean direction and concentration from the weighted statistics of their features.

Together, these fitted parameters capture a part's typical appearance and its variation. How much a match to that part contributes to a class score is a separate question, answered by the mixture weight. This separation lets us extend our class-specific classifier by replacing its separate prototype banks with a single bank available to all classes. For example, two species that use the same wing pattern as evidence can share the prototype's mean direction and concentration while learning their own weights for it.

Our contributions are threefold:
\begin{itemize}
    \item We introduce spherical part prototypes that capture both a typical appearance and its variation through a mean direction and a prototype-specific concentration.
    \item We develop a class-specific classifier based on mixtures of these distributions and extend it to a shared pool, so different classes can use the same part prototype with their own weights.
    \item On CUB-200-2011~\cite{wah2011cub} and AP-10K~\cite{yu2021ap10k}, both the class-specific and shared-pool variants of \method{} achieve better overall explanation metrics than their respective baselines~\cite{zhu2025nonparampartprototypes,huang2023evalprotopnet,nauta2023pipnet}, while remaining competitive in classification accuracy.
The class-specific model also achieves competitive accuracy on Stanford Dogs~\cite{KhoslaYaoJayadevaprakashFeiFei_FGVC2011} and Stanford Cars~\cite{krause2013cars}.
\end{itemize}

\section{Related Work}
\label{sec:related}

\textbf{Foundation Vision Models.}
Foundation models learn visual representations through approaches such as self-supervised contrastive learning~\cite{chen2020simclr} and language supervision for zero-shot recognition~\cite{radford2021clip}. Some of these representations are useful for part-based classification because they preserve information about the spatial layout of the image. In self-supervised Vision Transformers~\cite{dosovitskiy2020image}, for example, patch tokens often align with object regions and parts without pixel-level supervision~\cite{caron2021dino,oquab2024dinov2,simeoni2025dinov3}. This alignment has been used for unsupervised object discovery~\cite{simeoni2021lost}, and it also lets a prototype classifier look for a part by comparing individual patch features with a prototype. Such comparisons often use cosine similarity, which depends on directions rather than magnitudes. We therefore model part variation with distributions on the unit sphere.

\textbf{Part-Prototype Networks.}
A prototype network uses the similarity between a patch and a prototype as evidence for a class. ProtoPNet aggregates these similarities within a separate prototype bank for each class~\cite{chen2019thislookslikethat}, whereas ProtoPool and PIP-Net let several classes draw evidence from the same prototype~\cite{rymarczyk2022protopool,nauta2023pipnet}. Sharing changes which classes can use a match, while other variants make spatial matching more flexible~\cite{donnelly2022deformableprotopnet,ma2024protovit} or adapt prototype networks to transformer architectures~\cite{xue2024protopformer,ma2024protovit}. Once matches are available, models can also differ in how they use them to make a prediction. Some use tree-structured decisions~\cite{nauta2021prototree}, while ProtoKNN uses $k$-nearest-neighbor matching~\cite{ukai2023protoknn}. Alongside these changes to the classifiers, work on explanation quality has introduced multiple visualizations and feature-disambiguated prototypes~\cite{ma2023protoconcepts,pach2025lucidppn}, as well as diagnostics for localization, stability, and spatial misalignment~\cite{huang2023evalprotopnet,sacha2024misalignment}.

These methods differ in how they use prototype matches, but the matches themselves also depend on how patches are assigned and what each prototype can represent. NPPP focuses on assignment, using optimal transport (OT) solved by Sinkhorn iterations to structure patch-to-prototype assignments and discourage collapse onto a few recurring patterns~\cite{zhu2025nonparampartprototypes}. MGProto changes the representation by fitting a mixture of Gaussians whose components can have different means but share a fixed isotropic covariance, giving them all the same spread~\cite{wang2025mgproto}. \method{} follows NPPP in using OT to guide prototype discovery. Unlike MGProto, it gives each spherical prototype its own fitted concentration, so spread can vary across prototypes in both the class-specific and shared-pool models.

\section{Method} \label{sec:method}

\begin{figure*}[t]
  \centering
  \includegraphics[width=0.84\textwidth,height=0.65\textheight,keepaspectratio]{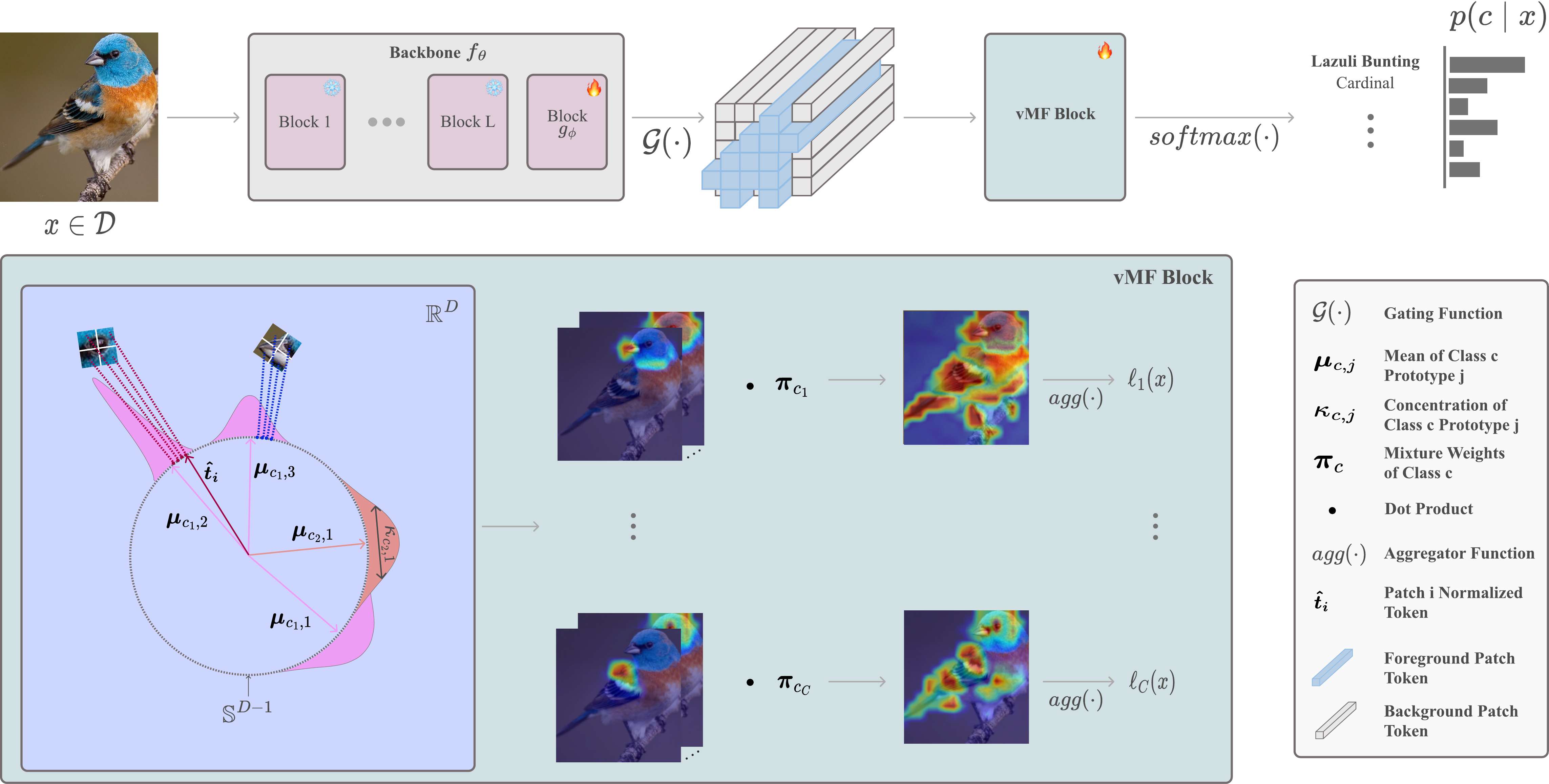}
  \caption{\textbf{\method{} framework.} \textbf{Top:} A frozen backbone and a trainable block extract patch features. Attention and PCA select foreground patches for classification. \textbf{Bottom:} vMF prototypes score normalized patches, and mixture weights determine their contributions to each class.}
  \label{fig:method}
\end{figure*}

We propose \method{}, a framework for fine-grained, explainable image classification that represents semantic parts as distributions on the hypersphere rather than as point prototypes. As illustrated in \cref{fig:method}, we adapt the features of a frozen vision transformer with a single trainable block and normalize the resulting patch tokens to unit length. On this sphere, the class-specific model represents each class as a mixture of von Mises--Fisher (vMF) distributions, one for each part prototype. Each prototype's mean direction describes a typical feature, while its fitted concentration $\kappa$ controls the spread around that direction. This lets the model represent a rigid part with a tighter distribution and a more deformable part with a broader one. A label-free foreground gate restricts classification and fitting to foreground patches. During discovery, we use entropic optimal transport (OT) to structure the patch-to-prototype assignments and encourage prototypes to cover different visual patterns rather than collapse onto a few.

\subsection{Overview}
\label{subsec:overview}

\textbf{Problem Formulation and Notation.} Let $\mathcal{D}=\{(x,y)\}$ be a dataset of images $x$ with class labels $y\in\{1,\ldots,C\}$. For each image $x$, the frozen backbone $f_\theta$ produces $N$ patch tokens $T^{\mathrm{froz}}$, which the trainable block $g_\phi$ maps to $T=\{\boldsymbol{t}_i\}_{i=1}^N$, with $\boldsymbol{t}_i\in\mathbb{R}^D$. To compare these adapted tokens with spherical prototypes, we normalize each one as $\hat{\boldsymbol{t}}_i=\boldsymbol{t}_i/\lVert\boldsymbol{t}_i\rVert_2\in\mathbb{S}^{D-1}$. Of the resulting prototype matches, only those from foreground patches contribute to the class scores. To select those patches, we compute a binary mask $w(x)\in\{0,1\}^N$ from the frozen backbone without labels. The classifier then aggregates the selected matches into image-level scores.

\textbf{Prototype Representation.} To score the selected tokens, the class-specific model gives each class $c$ a bank of $J$ prototypes, one per vMF component. The vMF component for prototype $j$ is centered at the mean direction $\boldsymbol{\mu}_{c,j}\in\mathbb{S}^{D-1}$, and its concentration $\kappa_{c,j}>0$ controls the spread around that direction. The larger the concentration, the tighter the distribution. These parameters describe the prototype itself, while a separate mixture weight $\pi_{c,j}\in[0,1]$ determines how much a match to it contributes to class $c$, with $\sum_{j=1}^{J}\pi_{c,j}=1$. Because the weights are separate from the distribution parameters, the shared-pool extension lets all classes draw from one bank of distributions while keeping their own mixture weights over that bank.

\textbf{Discovery and Fitting.} During prototype discovery, we use entropic optimal transport (OT) to structure patch-to-prototype assignments and encourage the prototypes to cover different visual patterns. We then transition to fitting each prototype's mean direction and concentration from weighted statistics of its associated foreground patches. Once this transition is complete, the feature block and class weights continue to train, but the distribution parameters come from the fit rather than from independent optimization.

\subsection{Backbone and Foreground Gating}
\label{subsec:backbone_gating}

We keep the pretrained DINO ViT frozen and append a single trainable transformer block $g_\phi$~\cite{bafghi2024parameter}. The added block adapts the features we match to prototypes, while the foreground mask $w(x)$ depends only on the frozen backbone. To build that mask without labels, we use the backbone's class-token attention to identify a candidate region, then refine it with principal component analysis (PCA) on the frozen patch tokens. We use the normalized adapted tokens $\hat{\boldsymbol{t}}_i$ at the selected locations for classification and prototype fitting. Appendix~\ref{subsec:supp_foreground_extraction} details how we compute the mask.

\subsection{Distributional Prototypes and Class Scoring}
\label{subsec:vmf_prototypes}

A point prototype gives a reference direction for a part, but it does not describe how much the part's features vary around that direction. We therefore represent each class $c$ with $J$ von Mises--Fisher (vMF) components on the unit sphere. The unit mean direction $\boldsymbol{\mu}_{c,j}$ describes a typical feature for prototype $j$, and its concentration $\kappa_{c,j}>0$ controls how tightly the features cluster around it. Because each prototype has its own concentration, it can describe a different amount of variation. For a patch token $\hat{\boldsymbol{t}}$ of unit length, the density of a component with parameters $\boldsymbol{\mu}$ and $\kappa$ is
\begin{equation}
f_{\mathrm{vMF}}(\hat{\boldsymbol{t}};\boldsymbol{\mu},\kappa)
=C_D(\kappa)\exp\!\bigl(\kappa\boldsymbol{\mu}^{\top}\hat{\boldsymbol{t}}\bigr),
\label{eq:vmf_pdf}
\end{equation}
where $C_D(\kappa)=\kappa^{D/2-1}/\bigl((2\pi)^{D/2}I_{D/2-1}(\kappa)\bigr)$ is the normalizing constant and $I_\nu$ is the modified Bessel function of the first kind.

\textbf{Patch evidence.} We use the component density to score a foreground patch against a prototype. For prototype $j$ in class $c$, let $a_{i,c,j}$ denote its log density evaluated at patch token $\hat{\boldsymbol{t}}_i$. We compare these scores within each candidate class, so a patch selects one prototype from each class. In class $c$, the winner has index
\begin{equation}
j^\star(i,c)=\arg\max_{1\leq j\leq J} a_{i,c,j}.
\label{eq:patch_winner}
\end{equation}
Only after choosing the winner do we apply the class's mixture weights $\pi_{c,j}\geq 0$, which satisfy $\sum_{j=1}^{J}\pi_{c,j}=1$. With $j^\star=j^\star(i,c)$, the evidence is
\begin{equation}
e_{i,c}=\pi_{c,j^\star}a_{i,c,j^\star}.
\label{eq:patch_evidence}
\end{equation}

\textbf{Image logits.} The winner rule gives each foreground patch one evidence value per class. To score the image, we use the evidence from every patch in the foreground set $\mathcal{F}(x)=\{i:w_i(x)=1\}$. When this set is nonempty, we average that evidence to obtain the logit for class $c$:
\begin{equation}
\ell_c(x)=\frac{1}{|\mathcal{F}(x)|}\sum_{i\in\mathcal{F}(x)}e_{i,c}.
\label{eq:class_logit}
\end{equation}
By averaging rather than summing the evidence, we keep the number of foreground patches from directly scaling the logit. We then apply a softmax across these logits to obtain the class probabilities:
\begin{equation}
p(c\mid x)=\frac{\exp(\ell_c(x))}{\sum_{c'=1}^{C}\exp(\ell_{c'}(x))}.
\label{eq:softmax}
\end{equation}
The next subsection explains how we discover these prototypes and fit their distributions.

\subsection{Prototype Discovery and Statistical Fitting}
\label{subsec:assignment_learning}

\textbf{Prototype discovery.} At prediction time, each patch selects a winner independently within each class. During discovery, the same rule can send most patches to a few prototypes or let several prototypes learn the same pattern. We therefore assign patches jointly within each training batch using entropic optimal transport (OT), following Zhu et al.~\cite{zhu2025nonparampartprototypes}. For the $M$ foreground patches from images of class $c$, indexed by $\mathcal I_c$, we form the score matrix
\begin{equation}
S_{ij}=a_{i,c,j}
=\log C_D(\kappa_{c,j})
+\kappa_{c,j}\boldsymbol{\mu}_{c,j}^{\top}\hat{\boldsymbol t}_i.
\label{eq:ot_score}
\end{equation}
We can then compute the transport plan as
\begin{equation}
\begin{aligned}
Q^*&=\arg\max_{Q\geq 0}
\left\{\langle Q,S\rangle+\varepsilon H(Q)\right\},\\
\text{subject to}\qquad
Q\mathbf 1_J&=\tfrac{1}{M}\mathbf 1_M,\qquad
Q^\top\mathbf 1_M=\boldsymbol\pi_c,
\end{aligned}
\label{eq:entropic_ot}
\end{equation}
where $\varepsilon>0$ weights the entropy $H(Q)$. The constraints on the transport plan give each patch the same mass and set the total mass received by each prototype according to the class mixture weights. We solve for the soft plan using Sinkhorn iterations~\cite{cuturi2013sinkhorn,peyre2019computationalot}, then take the largest entry in each row as the hard training target:
\begin{equation}
j^{\mathrm{OT}}(i,c)=\arg\max_j Q^*_{ij},
\qquad
z_{i,c,j}=\mathbb I[j=j^{\mathrm{OT}}(i,c)].
\label{eq:ot_harden}
\end{equation}

During warm-up, we first use these assignments to initialize the prototype directions while keeping the backbone and the added block frozen, holding $\kappa$ at its initial value, and using uniform mixture weights. With momentum $\gamma\in[0,1)$ and $\operatorname{norm}(\boldsymbol v)=\boldsymbol v/\|\boldsymbol v\|_2$, we update each direction toward its assigned patches:
\begin{equation}
\boldsymbol\mu_{c,j}\leftarrow
\operatorname{norm}\!\left(
\gamma\boldsymbol\mu_{c,j}
+(1-\gamma)\sum_{i\in\mathcal I_c}z_{i,c,j}\hat{\boldsymbol t}_i
\right).
\label{eq:mu_momentum}
\end{equation}

After warm-up, we train the added block, prototype directions, concentrations, and mixture weights by gradient descent. To carry the OT supervision into this stage, we use the hard assignments as targets for patch-to-prototype distillation (PPD). Following Zhu et al.~\cite{zhu2025nonparampartprototypes}, PPD compares these targets with a posterior over the prototypes of class $c$, obtained from the model's scores:
\begin{equation}
p(j\mid\hat{\boldsymbol t}_i, c)=
\frac{\exp(a_{i,c,j})}
{\sum_{j'=1}^J\exp(a_{i,c,j'})}.
\label{eq:posterior_k}
\end{equation}
We then train these probabilities with cross-entropy against the hard assignments.
\begin{equation}
\mathcal L_{\mathrm{PPD}}=
-\frac{1}{M}\sum_{i\in\mathcal{I}_c}\sum_{j=1}^J z_{i,c,j}\,\log p(j \mid \boldsymbol{\hat t}_i, c).
\label{eq:distill_loss}
\end{equation}

While PPD supervises patch assignments, LogDet acts directly on the prototype distributions by penalizing their overlap. For two vMF densities $f_{c,j}$ and $f_{c,k}$ in the same class, this overlap is their normalized $L^2$ inner product (see the appendix for its closed form):
\begin{align}
K^{(c)}_{jk}
&=\frac{\langle f_{c,j},f_{c,k}\rangle_{L^2}}
{\|f_{c,j}\|_{L^2}\|f_{c,k}\|_{L^2}},
\label{eq:vmf_kernel}\\
\mathcal L_{\mathrm{LogDet}}
&=-\frac{1}{C}\sum_{c=1}^{C}
\log\det\!\left(K^{(c)}+\delta I\right).
\label{eq:logdet}
\end{align}
Here $C$ is the number of foreground classes. The diagonal term $\delta I$ with $\delta>0$ stabilizes the determinant, while nearly identical prototype distributions still yield similar rows in $K^{(c)}$ and incur a larger penalty. Together with cross-entropy on the image logits, these losses give the discovery objective
\begin{equation}
\mathcal L_{\mathrm{disc}}
=\mathcal L_{\mathrm{CE}}
+\lambda_{\mathrm{PPD}}\mathcal L_{\mathrm{PPD}}
+\lambda_{\mathrm{LogDet}}\mathcal L_{\mathrm{LogDet}}.
\label{eq:total_loss}
\end{equation}

\textbf{Statistical fitting.} After discovery, we assign each foreground patch to the highest-scoring prototype of its labeled class and fit each prototype to all of its assigned patches. Within each image, patches that align more closely with the discovered direction receive more weight, but we normalize these weights so that every image with a match contributes the same total weight to that prototype's fit. For prototype $j$ in class $c$, we denote the set of assigned support image and patch pairs $(x,i)$ by $\mathcal P_{c,j}$ and their resulting weights by $\omega_{x,i,c,j}>0$. We use these weights to compute the mean of the assigned features and its length:
\begin{equation}
\bar{\boldsymbol t}_{c,j}=
\frac{\sum_{(x,i)\in\mathcal P_{c,j}}\omega_{x,i,c,j}\hat{\boldsymbol t}_i(x)}
{\sum_{(x,i)\in\mathcal P_{c,j}}\omega_{x,i,c,j}},
\qquad
\rho_{c,j}=\|\bar{\boldsymbol t}_{c,j}\|_2.
\label{eq:fit_patch_moment}
\end{equation}
The weighted mean supplies both fitted parameters. We take its direction as the prototype's mean direction and, since its length $\rho_{c,j}$ reflects how closely the assigned features align, choose the concentration to match that length:
\begin{equation}
\begin{gathered}
\boldsymbol\mu^{\mathrm{fit}}_{c,j}
=\frac{\bar{\boldsymbol t}_{c,j}}{\rho_{c,j}},
\qquad
A_D(\kappa^{\mathrm{fit}}_{c,j})=\rho_{c,j},\\
A_D(\kappa)=\frac{I_{D/2}(\kappa)}{I_{D/2-1}(\kappa)},
\end{gathered}
\label{eq:fit_vmf_moments}
\end{equation}
where $I_\nu$ is the modified Bessel function of the first kind.

These moments depend on the features produced by the added block, so we recompute them as the block changes during adaptation, while keeping the assignments and weights from the end of discovery. As adaptation progresses, we give the resulting fit more weight by interpolating between the freely learned and fitted vMF natural parameters:
\begin{equation}
\boldsymbol\eta_{c,j}
=(1-\alpha)\kappa^{\mathrm{free}}_{c,j}\boldsymbol\mu^{\mathrm{free}}_{c,j}
+\alpha\kappa^{\mathrm{fit}}_{c,j}\boldsymbol\mu^{\mathrm{fit}}_{c,j},
\qquad \alpha\in[0,1].
\label{eq:fit_transition}
\end{equation}
The length of $\boldsymbol\eta_{c,j}$ gives the concentration, and its direction gives the mean direction. Once $\alpha=1$, supported prototypes get both parameters from the fit. The mixture weights and the added block, by contrast, continue to train, with gradients reaching the block through both the scored patches and the support moments.

During adaptation, we continue to train with cross-entropy while using the model at the end of discovery as a fixed reference for prediction distillation. Let $q_c(x)$ be the reference model's class probabilities and $p_c(x)$ the softmax of the current training logits. We compare the two distributions with the prediction loss
\begin{equation}
\mathcal L_{\mathrm{pred}}
=\mathbb E_x\!\left[
\sum_{c=1}^{C}q_c(x)\log\frac{q_c(x)}{p_c(x)}
\right],\\
\end{equation}
Minimizing this loss keeps the current class probabilities close to those from discovery.

We also supervise where each prototype responds most strongly. For prototype $j$ of the image's labeled class $c$, let $b_i=\boldsymbol\mu_{c,j}^{\top}\hat{\boldsymbol t}_i(x)$ be its cosine score at location $i$. The peak penalty uses these cosine scores to favor a target location $r$ over the other locations
\begin{equation}
\begin{aligned}
&\Phi(b,r;\kappa_{c,j})\\
&\quad=\frac{1}{\kappa_{c,j}}
\log\!\left(
1+\sum_{i\ne r}
\exp\!\left[\kappa_{c,j}(b_i-b_r+\Delta)\right]
\right).
\end{aligned}
\end{equation}
The margin $\Delta>0$ encourages the target location to score above the other locations by a clear gap. To retain the reference peak, we score the discovery features with the direction fitted at the end of discovery and take the highest-scoring location as $r$. Using that reference gives the peak loss
\begin{equation}
\mathcal L_{\mathrm{peak}}
=\mathbb E_{x,j}\!\left[
\Phi(b,r;\kappa_{c,j})\right].
\end{equation}

For each image, we compute the score map $b$ separately for the clean image and a noise-perturbed version. We apply the penalty to both maps with the same fixed reference location $r$ and average the resulting penalties over images, their two views, and the prototypes of each image's labeled class.

The prediction and peak losses complement cross-entropy and the remaining OT supervision in the fitting objective:
\begin{equation}
\begin{aligned}
\mathcal L_{\mathrm{fit}}
={}&\mathcal L_{\mathrm{CE}}
+\lambda_{\mathrm{PPD}}^{\mathrm{fit}}\mathcal L_{\mathrm{PPD}}\\
&+\lambda_{\mathrm{pred}}\mathcal L_{\mathrm{pred}} +\lambda_{\mathrm{peak}}\mathcal L_{\mathrm{peak}}.
\end{aligned}
\end{equation}
As the fitted parameters take over, we phase out PPD by reducing its weight in this objective, which omits LogDet throughout adaptation. For the final refit, we hold the learned features and mixture weights fixed and alternate between updating the hard assignments under the current scores and fitting the distributions to those assignments on the training set.

Beyond the class-specific method we also make an extension to a single pool of fitted vMF prototypes shared by all classes. Where each class learns its own weights over the pool, so the same prototype can contribute across different classes. The appendix describes how we adapt scoring and training to the shared pool.

\section{Experiments}\label{sec:exp}

\subsection{Experimental Setup}
\label{subsec:exp_setup}
\begin{table*}[!t]
  \centering
  \setlength{\tabcolsep}{2.9pt}
  \renewcommand{\arraystretch}{0.97}
  \setlength{\aboverulesep}{0.3ex}
  \setlength{\belowrulesep}{0.4ex}
  \caption{\textbf{Class-specific results on CUB.} All metrics are percentages, bold and underline mark the best and second-best values.}
  \label{tab:cub_main}
  \small
  \begin{tabular}{lc|cccc|cccc|cccc|cccc}
    \toprule
    Method & $J$ & \multicolumn{4}{c|}{\textbf{DINOv2 ViT-B/14}} & \multicolumn{4}{c|}{\textbf{DINOv2 ViT-S/14}} & \multicolumn{4}{c|}{\textbf{DINOv3 ViT-B/16}} & \multicolumn{4}{c}{\textbf{DINOv3 ViT-S/16}} \\
     & & Con. & Sta. & Dis. & Acc. & Con. & Sta. & Dis. & Acc. & Con. & Sta. & Dis. & Acc. & Con. & Sta. & Dis. & Acc. \\
    \midrule
ProtoPNet & 5 & 8.0 & 62.6 & 90.3 & 64.3 & 8.0 & 51.9 & 87.0 & 13.4 & 11.1 & 75.6 & 88.9 & 77.5 & 20.4 & 66.8 & 86.9 & 72.8 \\
ProtoPNet & 7 & 5.4 & 61.9 & 89.6 & 45.8 & 2.3 & 60.0 & 88.4 & 46.0 & 15.6 & 74.5 & 88.6 & 80.0 & 22.5 & 69.6 & 84.5 & 72.5 \\
\midrule
Def. ProtoPNet & 5 & 18.4 & 73.4 & 65.8 & 86.7 & 22.5 & 72.6 & 57.7 & 86.1 & 17.7 & 73.8 & 79.2 & 84.9 & 11.3 & 69.8 & 81.5 & 77.4 \\
Def. ProtoPNet & 7 & 28.8 & 77.7 & 53.5 & 87.9 & 16.5 & 66.7 & 61.0 & 86.8 & 27.2 & 77.3 & 72.3 & 86.6 & 12.9 & 70.8 & 78.6 & 82.2 \\
    \midrule
TesNet~\cite{wang2021tesnet} & 5 & 28.6 & 74.8 & 37.8 & 85.2 & 43.9 & 64.8 & 86.9 & 82.6 & 37.4 & 60.7 & 62.7 & 83.8 & 47.3 & 65.2 & 82.1 & 79.6 \\
TesNet & 7 & 42.0 & 76.9 & 38.8 & 86.8 & 50.7 & 61.3 & 76.2 & 81.3 & 39.3 & 65.6 & 43.7 & 83.5 & 39.8 & 64.2 & 74.9 & 80.7 \\
\midrule
EvalProtoPNet & 5 & 53.8 & 71.2 & 36.5 & 83.2 & 42.6 & 75.6 & 46.6 & 84.6 & 70.4 & 74.6 & 26.8 & 83.7 & 59.2 & 69.2 & 36.7 & 82.7 \\
EvalProtoPNet & 7 & 49.4 & 70.8 & 45.8 & 80.4 & 43.5 & 75.3 & 64.1 & 83.4 & 60.1 & 75.1 & 41.9 & 82.0 & 67.5 & 76.1 & 51.6 & 82.9 \\
\midrule
MGProto & 5 & 16.7 & 68.9 & 53.2 & 83.7 & 1.9 & 52.4 & 27.5 & 72.4 & 35.5 & 61.4 & 52.6 & 86.0 & 7.0 & 50.8 & 20.8 & 77.6 \\
MGProto & 7 & 8.3 & 63.1 & 12.7 & 79.6 & 4.5 & 58.9 & 11.9 & 74.2 & 17.6 & 56.8 & 11.4 & 85.1 & 9.8 & 55.1 & 6.3 & 75.8 \\
\midrule
NPPP & 5 & 58.5 & 81.0 & 81.0 & \underline{90.8} & 10.2 & \underline{84.3} & 57.0 & 84.1 & 0.6 & \textbf{90.7} & 48.0 & 83.9 & 0.6 & \textbf{90.0} & 42.5 & 57.9 \\
NPPP & 7 & 61.6 & 82.3 & 77.7 & \textbf{90.9} & 14.6 & \textbf{85.1} & 57.4 & 85.1 & 0.3 & \underline{90.6} & 48.8 & 86.0 & 0.9 & \underline{89.4} & 43.0 & 63.0 \\
    \midrule
    \method{} & 5 & \textbf{80.2} & \underline{84.3} & \textbf{95.9} & 90.5 & \textbf{78.9} & 78.7 & \textbf{95.8} & \textbf{88.6} & \textbf{81.9} & 85.6 & \textbf{95.1} & \textbf{90.3} & \underline{68.0} & 80.2 & \textbf{94.8} & \textbf{86.9} \\
    \method{} & 7 & \underline{77.8} & \textbf{84.8} & \underline{94.7} & 90.7 & \underline{76.3} & 80.2 & \underline{94.5} & \underline{88.5} & \underline{77.9} & 85.1 & \underline{93.5} & \underline{90.2} & \textbf{70.1} & 80.7 & \underline{93.1} & \underline{86.1} \\
    \bottomrule
  \end{tabular}
\end{table*}

\textbf{Datasets.}
We use CUB-200-2011~\cite{wah2011cub} as our main benchmark for classification and part explanations.
AP-10K~\cite{yu2021ap10k} provides a second benchmark for both, while Stanford Dogs~\cite{KhoslaYaoJayadevaprakashFeiFei_FGVC2011} and Stanford Cars~\cite{krause2013cars} extend the accuracy evaluation.
We train on the full official training split of each dataset and use part annotations only to evaluate explanations.

\textbf{Metrics.}
We report top-1 accuracy (\emph{Acc.}) on all four datasets.
On CUB and AP-10K, we also report prototype explanation metrics, including consistency (\emph{Con.}), which measures how often a prototype matches the same annotated part across images, and stability (\emph{Sta.}), which measures whether those matches persist under small input perturbations~\cite{huang2023evalprotopnet}.
Distinctiveness (\emph{Dis.}) measures how distinct the responses of prototypes within a class are, with lower scores when prototypes highlight the same regions~\cite{zhu2025nonparampartprototypes}.

\textbf{Backbones and prototype budgets.}
Across datasets, we use DINOv2 ViT-B/14~\cite{oquab2024dinov2} and DINOv3 ViT-B/16~\cite{simeoni2025dinov3}.
On CUB, we also evaluate class-specific prototypes with the corresponding Small backbones, ViT-S/14 and ViT-S/16.
We use $J=5$ or $7$ prototypes per class in the class-specific setting.
When classes share a pool instead, we use $R$ prototypes in total, with matched pool sizes of 512 or 1,000 on CUB and 125 or 250 on AP-10K.

\textbf{Baselines and training.}
On AP-10K, we compare class-specific \method{} with NPPP~\cite{zhu2025nonparampartprototypes} and EvalProtoPNet~\cite{huang2023evalprotopnet}, while on CUB we include additional baselines~\cite{chen2019thislookslikethat,donnelly2022deformableprotopnet,wang2021tesnet,wang2025mgproto}.
For shared prototypes, we compare with PIP-Net~\cite{nauta2023pipnet} and ProtoPool~\cite{rymarczyk2022protopool} on CUB and AP-10K.
In the main comparisons, we match the DINO backbone, train each method for 25 epochs, and use the same evaluation protocol, while each baseline keeps its own foreground handling.
For \method{}, the 25 epochs cover the OT warm-up, prototype discovery, and statistical fitting described in \cref{subsec:assignment_learning}.
We use Adam~\cite{kingma2015adam} for gradient-based updates and list the training hyperparameters in the appendix.

\begin{table}[!t]
  \centering
  \setlength{\tabcolsep}{1.55pt}
  \renewcommand{\arraystretch}{0.97}
  \setlength{\aboverulesep}{0.3ex}
  \setlength{\belowrulesep}{0.4ex}
  \caption{\textbf{Class-specific results on AP-10K.} All metrics are percentages, bold and underline mark the best and second-best values.}
  \label{tab:ap10k_main}
  \small
  \begin{tabular}{@{}lc|cccc|cccc@{}}
    \toprule
    Method & $J$ & \multicolumn{4}{c|}{\textbf{DINOv2 ViT-B/14}} & \multicolumn{4}{c}{\textbf{DINOv3 ViT-B/16}} \\
     & & Con. & Sta. & Dis. & Acc. & Con. & Sta. & Dis. & Acc. \\
    \midrule
    
EvalProtoPNet & 5 & \underline{95.9} & 82.0 & 15.1 & 97.3 & 95.9 & 88.0 & 14.6 & 95.1 \\
EvalProtoPNet & 7 & \textbf{96.3} & 82.1 & 33.9 & 97.2 & 96.4 & 86.2 & 40.3 & 93.3 \\
\midrule
NPPP & 5 & 3.0 & 79.4 & 84.1 & \underline{97.6} & \underline{96.5} & \underline{89.8} & 46.3 & \textbf{98.0} \\
NPPP & 7 & 2.6 & 78.5 & 80.8 & \textbf{97.7} & \textbf{97.3} & \textbf{90.6} & 48.8 & \underline{97.9} \\

    \midrule
    \method{} & 5 & 87.6 & \textbf{83.0} & \textbf{91.9} & \underline{97.6} & 95.6 & 87.3 & \textbf{89.4} & \textbf{98.0} \\
    \method{} & 7 & 86.0 & \underline{82.9} & \underline{90.3} & 97.5 & 94.6 & 87.1 & \underline{87.3} & 97.6 \\
    \bottomrule
  \end{tabular}
\end{table}

\begin{table*}[!t]
  \centering
  \setlength{\tabcolsep}{2.5pt}
  \renewcommand{\arraystretch}{0.97}
  \setlength{\aboverulesep}{0.3ex}
  \setlength{\belowrulesep}{0.4ex}
  \caption{\textbf{Shared-pool results on CUB and AP-10K.} Explanation metrics use the five highest-weighted prototypes per class.}
  \label{tab:shared_main}
  \small
  \begin{tabular}{l|cc|cccc|cccc|cccc|cccc}
    \toprule
    Method & \multicolumn{2}{c|}{\textbf{Pool size} $R$} & \multicolumn{8}{c|}{\textbf{CUB}} & \multicolumn{8}{c}{\textbf{AP-10K}} \\
    \cmidrule(lr){2-3}\cmidrule(lr){4-11}\cmidrule(lr){12-19}
     & \textbf{CUB} & \textbf{AP-10K} & \multicolumn{4}{c|}{\textbf{DINOv2 ViT-B/14}} & \multicolumn{4}{c|}{\textbf{DINOv3 ViT-B/16}} & \multicolumn{4}{c|}{\textbf{DINOv2 ViT-B/14}} & \multicolumn{4}{c}{\textbf{DINOv3 ViT-B/16}} \\
     & & & Con. & Sta. & Dis. & Acc. & Con. & Sta. & Dis. & Acc. & Con. & Sta. & Dis. & Acc. & Con. & Sta. & Dis. & Acc. \\
    \midrule
    PIP-Net & 512 & 125 & \underline{70.3} & 80.0 & 83.1 & \underline{90.7} & 68.2 & 81.4 & 85.2 & \textbf{89.8} & \underline{81.6} & \underline{86.9} & 64.6 & \textbf{97.9} & 60.4 & \underline{91.3} & 83.9 & \underline{97.2} \\
    PIP-Net & 1,000 & 250 & 68.3 & 81.7 & 84.3 & \textbf{91.4} & \underline{72.9} & 82.6 & 83.5 & 88.5 & \textbf{89.2} & \textbf{88.8} & 74.4 & \underline{97.6} & 73.6 & \textbf{92.5} & 82.0 & \textbf{97.7} \\
    \midrule
    ProtoPool & 512 & 125 & 30.9 & 71.0 & 87.9 & 42.2 & 8.6 & 69.4 & 83.4 & 46.5 & 45.2 & 63.4 & \underline{85.9} & 96.6 & 49.2 & 72.2 & 86.8 & 93.8 \\
    ProtoPool & 1,000 & 250 & 50.7 & 76.4 & 78.6 & 60.7 & 8.5 & 67.3 & 84.7 & 52.6 & 53.2 & 64.5 & 85.4 & 95.2 & 52.0 & 72.8 & 87.3 & 94.6 \\
    \midrule
    \method{} & 512 & 125 & 65.9 & \underline{82.4} & \underline{92.8} & 86.7 & 72.8 & \underline{83.4} & \underline{91.7} & 87.6 & 55.2 & 77.8 & 82.6 & 95.8 & \underline{82.8} & 77.1 & \underline{87.8} & 97.1 \\
    \method{} & 1,000 & 250 & \textbf{71.1} & \textbf{84.4} & \textbf{93.3} & 88.2 & \textbf{75.3} & \textbf{86.0} & \textbf{93.2} & \underline{88.8} & 80.8 & 80.3 & \textbf{91.5} & 95.8 & \textbf{94.8} & 83.7 & \textbf{88.0} & 97.1 \\
    \bottomrule
  \end{tabular}
\end{table*}

\subsection{Main Results}
\label{subsec:exp_main_results}

On CUB, class-specific \method{} produces more consistent and less redundant explanations across the tested backbones and prototype budgets while remaining competitive in accuracy (\cref{tab:cub_main}). Its prototypes return to the same parts across images, while different prototypes highlight different regions within an image. AP-10K shows why these properties matter together. Models with similar accuracy and consistency can still differ substantially in how much evidence their prototypes repeat, and \method{} gives more distinct responses in every matched comparison (\cref{tab:ap10k_main}). These results support modeling the variation of a part within a prototype while keeping the evidence from different prototypes complementary.

Sharing prototypes carries the same design into a setting where several classes can use the same parts. At matched backbones and pool sizes, \method{} gives less overlapping responses than PIP-Net on both datasets (\cref{tab:shared_main}). That separation matters alongside consistency and stability. These measures tell us whether a match holds up across images and perturbations, while distinctiveness tells us whether the other prototypes add evidence. On CUB, the three measures together favor \method{}'s explanations. On AP-10K, PIP-Net's more stable matches still repeat more of the same evidence. A shared pool can therefore reuse parts across classes and still give each class complementary evidence.

The appendix examines how the number of prototypes shown affects explanation consistency and how pruning the shared pool affects accuracy. It also reports classification accuracy on Stanford Dogs and Stanford Cars.

\begin{figure*}[!t]
  \centering
  \includegraphics[width=0.99\textwidth]{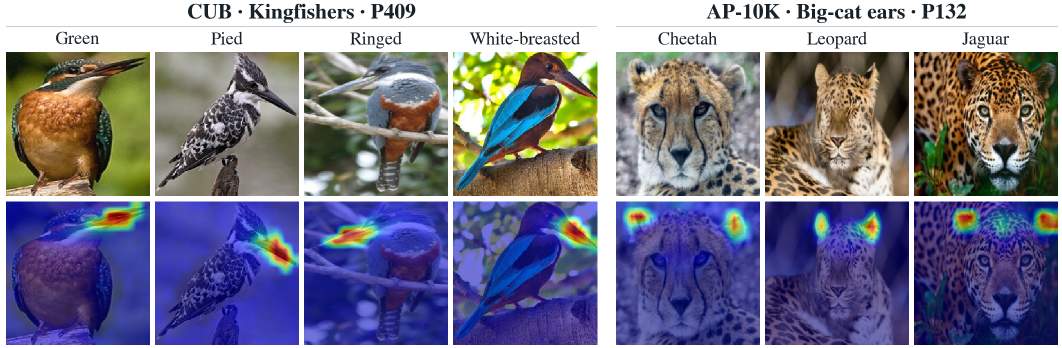}
  \caption{\textbf{Shared prototypes across species.} Test images (top) and activation maps (bottom) for CUB P409 (left) and AP-10K P132 (right). Each prototype ranks in the top five for every species shown in its panel.}
  \label{fig:cub_qual_shared}
\end{figure*}

\begin{figure*}[!t]
  \centering
  \includegraphics[width=0.99\textwidth]{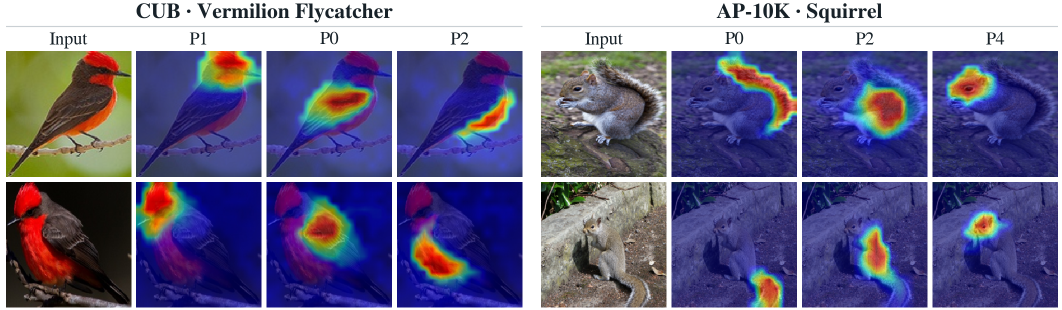}
  \caption{\textbf{Class-specific prototypes across images.} Each panel tracks three prototypes in fixed columns across two test images: Vermilion Flycatcher (CUB, left) and squirrel (AP-10K, right). Prototype IDs are local to each class.}
  \label{fig:cub_qual}
\end{figure*}

\subsection{Ablations}
\label{subsec:exp_ablations}

\begin{table}[!t]
  \centering
  \setlength{\tabcolsep}{2.0pt}
  \renewcommand{\arraystretch}{0.94}
  \caption{\textbf{Class-specific ablations on CUB.}}
  \label{tab:cub_ablation}
  \small
  \begin{tabular}{@{}p{0.51\linewidth}cccc@{}}
    \toprule
    Setting & Con. & Sta. & Dis. & Acc. \\
    \midrule
  \method{} (full) & 80.2 & 84.3 & 95.9 & 90.5 \\
    \addlinespace[1pt]
    \multicolumn{5}{l}{\textbf{Discovery losses}} \\
  CE only & 70.2 & 80.5 & 94.7 & 89.6 \\
  CE + PPD & 72.8 & 81.3 & 93.7 & 90.4 \\
  CE + LogDet & 62.5 & 78.0 & 96.0 & 89.1 \\
    \addlinespace[1pt]
    \multicolumn{5}{l}{\textbf{Training schedule}} \\
  Discovery only, 25 epochs & 66.8 & 77.7 & 96.2 & 90.6 \\
    \addlinespace[1pt]
    \multicolumn{5}{l}{\textbf{Concentration}} \\
  Fixed $\kappa=1$ & 55.0 & 77.2 & 95.8 & 85.6 \\
    \addlinespace[1pt]
    \multicolumn{5}{l}{\textbf{Fitting-stage losses}} \\
  CE only & 73.5 & 77.2 & 95.3 & 89.6 \\
  CE + prediction KL & 77.0 & 80.3 & 95.7 & 90.1 \\
  CE + reference peak & 77.5 & 78.7 & 95.6 & 89.5 \\
    \addlinespace[1pt]
    \multicolumn{5}{l}{\textbf{Distribution family}} \\
  Isotropic Gaussian\newline & 74.4 & 82.4 & 93.8 & 90.5 \\
    \bottomrule
  \end{tabular}
\end{table}

To understand what contributes to the explanation gains, we start by removing the fitting stage. On CUB, we compare \method{} with a model that continues discovery for the same training budget (\cref{tab:cub_ablation}). Fitting makes part matches more consistent across images and more stable under perturbations, with little change in accuracy or distinctiveness. These more reliable matches are accompanied by a closer fit to the assigned foreground features on the test set. Using each checkpoint's own assignments, the median angle between prototype directions and the empirical mean directions of their assigned features falls from \(26.3^\circ\) to \(6.9^\circ\). The concentration estimates also track the spread of these features more closely: across prototypes, the Spearman correlation between \(\kappa\) and empirical directional variance goes from \(-0.21\) to \(-0.8\), so higher concentrations correspond more consistently to tighter feature groups.

The loss ablations help explain why discovery and fitting use different supervision. During discovery, adding PPD to cross-entropy improves consistency, while adding LogDet instead makes prototype responses more distinct. Together they give the strongest consistency and stability among the discovery objectives tested while keeping distinctiveness high. Fitting builds on that initialization, so its losses help preserve what the model predicts and where its prototypes respond. Prediction distillation improves stability and accuracy over cross-entropy alone, while reference-peak supervision gives the highest consistency among the reduced objectives. The full fitting objective brings these forms of supervision together and gives stronger consistency and stability than either reduced variant.

The loss ablations tell us how to train the prototypes, while the representation controls test what changes when we constrain or replace their distributions. The fixed-\(\kappa=1\) control is less accurate and less consistent than the full model (\cref{tab:cub_ablation}). The Gaussian comparison separates explanation quality from accuracy more directly. An isotropic Gaussian fitted to raw features matches the reported accuracy of \method{} on unit-normalized features but falls behind on all three explanation metrics. The two models can therefore classify equally well while differing in whether their evidence is repeatable, stable, and complementary. The appendix reports further ablations on foreground gating and offline augmentation.

\subsection{Qualitative Results}
\label{subsec:exp_qual}

The activation maps show the behavior we expect from our prototypes. Each prototype can keep matching the same part as its appearance changes. In the shared pool, that makes a single prototype useful to several species without losing its focus on a specific part, as illustrated by kingfisher beaks and big-cat ears (Fig.~\ref{fig:cub_qual_shared}). Within a class, the flycatcher and squirrel examples show why repeatability and distinctiveness matter together. As the pose changes, each prototype returns to the same part, while different prototypes continue to highlight different regions (Fig.~\ref{fig:cub_qual}).

The shared examples in \cref{fig:cub_qual_shared} use DINOv2-B with $R=1{,}000$ on CUB and DINOv3-B with $R=250$ on AP-10K. The class-specific examples in \cref{fig:cub_qual} use DINOv2-B with $J=5$.

\let\MainFloatBarrier\FloatBarrier
\let\FloatBarrier\par
\section{Conclusion}\label{sec:concl}
We presented \method{}, an interpretable classifier that represents parts with distributions rather than points. By fitting a von Mises--Fisher distribution to the foreground features assigned to each prototype, \method{} captures both what a part typically looks like and how much its appearance varies. Across the backbones and prototype budgets tested on CUB, \method{} produced more consistent and less redundant explanations than the class-specific baselines while remaining competitive in accuracy. Compared with prototype discovery alone, statistical fitting improved consistency and stability at nearly unchanged accuracy. The fitted directions and concentrations also agreed more closely with the centres and spread of their assigned features on held-out images.

The gains extended to AP-10K and carried over to the shared-pool model. Sharing lets different classes use the same fitted prototype as evidence. With backbones and pool sizes matched, \method{} produced more distinctive explanations than PIP-Net on both datasets while remaining competitive in accuracy. An intriguing next step is to use samples from the fitted distributions to condition generative models and visualize the range of appearances each prototype represents.

\let\FloatBarrier\MainFloatBarrier

{\small
  
  \bibliographystyle{ieeenat_fullname}
  \bibliography{references}

@String(PAMI  = {IEEE Trans. Pattern Anal. Mach. Intell.})

@String(CVPR  = {IEEE Conf. Comput. Vis. Pattern Recog.})

@String(ICCV  = {Int. Conf. Comput. Vis.})

@String(ECCV  = {Eur. Conf. Comput. Vis.})

@String(NeurIPS = {Adv. Neural Inform. Process. Syst.})

@String(ICML  = {Int. Conf. Mach. Learn.})

@String(ICLR  = {Int. Conf. Learn. Represent.})

@String(BMVC  = {Brit. Mach. Vis. Conf.})

@String(CVPRW = {IEEE Conf. Comput. Vis. Pattern Recog. Worksh.})

@String(AAAI  = {AAAI})

@String(IJCAI = {IJCAI})

@String(JMLR  = {J. Mach. Learn. Res.})

@String(TMLR  = {Trans. Mach. Learn Res.})

@String(ICCVW = {ICCV Workshops})

@inproceedings{zhu2025nonparampartprototypes,
  title     = {Interpretable Image Classification via Non-parametric Part Prototype Learning},
  author    = {Zhu, Zhijie and Fan, Lei and Pagnucco, Maurice and Song, Yang},
  booktitle = CVPR,
  year      = {2025},
  url       = {https://openaccess.thecvf.com/content/CVPR2025/papers/Zhu_Interpretable_Image_Classification_via_Non-parametric_Part_Prototype_Learning_CVPR_2025_paper.pdf}
}

@article{wang2025mgproto,
  author  = {Wang, Chong and Chen, Yuanhong and Liu, Fengbei and Liu, Yuyuan and McCarthy, Davis James and Frazer, Helen and Carneiro, Gustavo},
  title   = {Mixture of {G}aussian-Distributed Prototypes with Generative Modelling for Interpretable and Trustworthy Image Recognition},
  journal = PAMI,
  volume  = {47},
  number  = {8},
  pages   = {6974--6989},
  year    = {2025},
  url     = {https://arxiv.org/abs/2312.00092}
}

@inproceedings{chen2019thislookslikethat,
  title     = {This Looks Like That: Deep Learning for Interpretable Image Recognition},
  author    = {Chen, Chaofan and Li, Oscar and Tao, Chaofan and Barnett, Alina Jade and Su, Jonathan and Rudin, Cynthia},
  booktitle = NeurIPS,
  year      = {2019},
  url       = {https://proceedings.neurips.cc/paper/2019/hash/adf7ee2dcf142b0e11888e72b43fcb75-Abstract.html}
}

@inproceedings{donnelly2022deformableprotopnet,
  author    = {Donnelly, Jon and Barnett, Alina Jade and Chen, Chaofan},
  title     = {Deformable {ProtoPNet}: An Interpretable Image Classifier Using Deformable Prototypes},
  booktitle = CVPR,
  year      = {2022},
  pages     = {10265--10275},
  url       = {https://openaccess.thecvf.com/content/CVPR2022/html/Donnelly_Deformable_ProtoPNet_An_Interpretable_Image_Classifier_Using_Deformable_Prototypes_CVPR_2022_paper.html}
}

@inproceedings{wang2021tesnet,
  author    = {Wang, Jiaqi and Liu, Huafeng and Wang, Xinyue and Jing, Liping},
  title     = {Interpretable Image Recognition by Constructing Transparent Embedding Space},
  booktitle = ICCV,
  year      = {2021},
  pages     = {895--904},
  url       = {https://openaccess.thecvf.com/content/ICCV2021/html/Wang_Interpretable_Image_Recognition_by_Constructing_Transparent_Embedding_Space_ICCV_2021_paper.html}
}

@inproceedings{nauta2023pipnet,
  author    = {Nauta, Meike and Schl{\"o}tterer, J{\"o}rg and van Keulen, Maurice and Seifert, Christin},
  title     = {{PIP-Net}: Patch-Based Intuitive Prototypes for Interpretable Image Classification},
  booktitle = CVPR,
  year      = {2023},
  pages     = {2744--2753},
  url       = {https://openaccess.thecvf.com/content/CVPR2023/html/Nauta_PIP-Net_Patch-Based_Intuitive_Prototypes_for_Interpretable_Image_Classification_CVPR_2023_paper.html}
}

@inproceedings{ma2024protovit,
  author    = {Ma, Chiyu and Donnelly, Jon and Liu, Wenjun and Vosoughi, Soroush and Rudin, Cynthia and Chen, Chaofan},
  title     = {Interpretable Image Classification with Adaptive Prototype-Based Vision Transformers},
  booktitle = NeurIPS,
  volume    = {37},
  pages     = {41447--41493},
  year      = {2024},
  url       = {https://proceedings.neurips.cc/paper_files/paper/2024/hash/48dfc849640344e2d58df0b5bb78c33b-Abstract-Conference.html}
}

@inproceedings{ma2023protoconcepts,
  author    = {Ma, Chiyu and Zhao, Brandon and Chen, Chaofan and Rudin, Cynthia},
  title     = {This Looks Like Those: Illuminating Prototypical Concepts Using Multiple Visualizations},
  booktitle = NeurIPS,
  volume    = {36},
  pages     = {39212--39235},
  year      = {2023},
  url       = {https://proceedings.neurips.cc/paper_files/paper/2023/hash/7b76eea0c3683e440c3d362620f578cd-Abstract-Conference.html}
}

@inproceedings{pach2025lucidppn,
  author    = {Pach, Mateusz and Lewandowska, Koryna and Tabor, Jacek and Zieli{\'n}ski, Bartosz and Rymarczyk, Dawid},
  title     = {{LucidPPN}: Unambiguous Prototypical Parts Network for User-Centric Interpretable Computer Vision},
  booktitle = ICLR,
  year      = {2025},
  url       = {https://openreview.net/forum?id=BM9qfolt6p}
}

@inproceedings{sacha2024misalignment,
  author    = {Sacha, Miko{\l}aj and Jura, Bartosz and Rymarczyk, Dawid and Struski, {\L}ukasz and Tabor, Jacek and Zieli{\'n}ski, Bartosz},
  title     = {Interpretability Benchmark for Evaluating Spatial Misalignment of Prototypical Parts Explanations},
  booktitle = AAAI,
  volume    = {38},
  pages     = {21563--21573},
  year      = {2024},
  doi       = {10.1609/aaai.v38i19.30154},
  url       = {https://ojs.aaai.org/index.php/AAAI/article/view/30154}
}

@inproceedings{nauta2021prototree,
  author    = {Nauta, Meike and van Bree, Ron and Seifert, Christin},
  title     = {Neural Prototype Trees for Interpretable Fine-Grained Image Recognition},
  booktitle = CVPR,
  year      = {2021},
  pages     = {14933--14943},
  url       = {https://openaccess.thecvf.com/content/CVPR2021/html/Nauta_Neural_Prototype_Trees_for_Interpretable_Fine-Grained_Image_Recognition_CVPR_2021_paper.html}
}

@inproceedings{rymarczyk2022protopool,
  author    = {Rymarczyk, Dawid and Struski, {\L}ukasz and G{\'o}rszczak, Micha{\l} and Lewandowska, Koryna and Tabor, Jacek and Zieli{\'n}ski, Bartosz},
  title     = {Interpretable Image Classification with Differentiable Prototypes Assignment},
  booktitle = ECCV,
  year      = {2022},
  url       = {https://arxiv.org/abs/2112.02902}
}

@inproceedings{ukai2023protoknn,
  author    = {Ukai, Yuki and Hirakawa, Tsubasa and Yamashita, Takayoshi and Fujiyoshi, Hironobu},
  title     = {This Looks Like It Rather Than That: {ProtoKNN} for Similarity-Based Classifiers},
  booktitle = ICLR,
  year      = {2023},
  url       = {https://openreview.net/forum?id=lh-HRYxuoRr}
}

@inproceedings{caron2021dino,
  author    = {Caron, Mathilde and Touvron, Hugo and Misra, Ishan and J{\'e}gou, Herv{\'e} and Mairal, Julien and Bojanowski, Piotr and Joulin, Armand},
  title     = {Emerging Properties in Self-Supervised Vision Transformers},
  booktitle = ICCV,
  year      = {2021},
  pages     = {9650--9660},
  url       = {https://arxiv.org/abs/2104.14294}
}

@article{dosovitskiy2020image,
  title   = {An Image is Worth 16x16 Words: Transformers for Image Recognition at Scale},
  author  = {Dosovitskiy, Alexey and Beyer, Lucas and Kolesnikov, Alexander and Weissenborn, Dirk and Zhai, Xiaohua and Unterthiner, Thomas and Dehghani, Mostafa and Minderer, Matthias and Heigold, Georg and Gelly, Sylvain and Uszkoreit, Jakob and Houlsby, Neil},
  journal = {arXiv preprint arXiv:2010.11929},
  year    = {2020},
  url     = {https://arxiv.org/abs/2010.11929}
}

@inproceedings{bafghi2024parameter,
  title     = {Parameter Efficient Fine-tuning of Self-supervised {ViTs} without Catastrophic Forgetting},
  author    = {Bafghi, Reza Akbarian and Harilal, Nidhin and Monteleoni, Claire and Raissi, Maziar},
  booktitle = CVPRW,
  pages     = {3679--3684},
  year      = {2024},
  url       = {https://openaccess.thecvf.com/content/CVPR2024W/ELVM/html/Bafghi_Parameter_Efficient_Fine-tuning_of_Self-supervised_ViTs_without_Catastrophic_Forgetting_CVPRW_2024_paper.html}
}

@misc{simeoni2025dinov3,
  author       = {Sim{\'e}oni, Oriane and Vo, Huy V. and Seitzer, Maximilian and Baldassarre, Federico and Oquab, Maxime and Jose, Cijo and Khalidov, Vasil and Szafraniec, Marc and Yi, Seungeun and Ramamonjisoa, Micha{\"e}l and Massa, Francisco and Haziza, Daniel and Wehrstedt, Luca and Wang, Jianyuan and Darcet, Timoth{\'e}e and Moutakanni, Th{\'e}o and Sentana, Leonel and Roberts, Claire and Vedaldi, Andrea and Tolan, Jamie and Brandt, John and Couprie, Camille and Mairal, Julien and J{\'e}gou, Herv{\'e} and Labatut, Patrick and Bojanowski, Piotr},
  title        = {{DINOv3}},
  howpublished = {arXiv:2508.10104},
  year         = {2025},
  doi          = {10.48550/arXiv.2508.10104},
  url          = {https://arxiv.org/abs/2508.10104}
}

@inproceedings{simeoni2021lost,
  author    = {Sim{\'e}oni, Oriane and Puy, Gilles and Vo, Huy V. and Roburin, Simon and Gidaris, Spyros and Bursuc, Andrei and P{\'e}rez, Patrick and Marlet, Renaud and Ponce, Jean},
  title     = {{LOST}: Localizing Objects with Self-Supervised Transformers and no Labels},
  booktitle = BMVC,
  year      = {2021},
  url       = {https://arxiv.org/abs/2109.14279}
}

@inproceedings{radford2021clip,
  author    = {Radford, Alec and Kim, Jong Wook and Hallacy, Chris and Ramesh, Aditya and Goh, Gabriel and Agarwal, Sandhini and Sastry, Girish and Askell, Amanda and Mishkin, Pamela and Clark, Jack and Krueger, Gretchen and Sutskever, Ilya},
  title     = {Learning Transferable Visual Models From Natural Language Supervision},
  booktitle = ICML,
  year      = {2021},
  url       = {https://arxiv.org/abs/2103.00020}
}

@inproceedings{chen2020simclr,
  author    = {Chen, Ting and Kornblith, Simon and Norouzi, Mohammad and Hinton, Geoffrey},
  title     = {A Simple Framework for Contrastive Learning of Visual Representations},
  booktitle = ICML,
  year      = {2020},
  pages     = {1597--1607},
  url       = {https://arxiv.org/abs/2002.05709}
}

@inproceedings{kingma2015adam,
  author    = {Kingma, Diederik P. and Ba, Jimmy},
  title     = {Adam: A Method for Stochastic Optimization},
  booktitle = ICLR,
  year      = {2015},
  url       = {https://arxiv.org/abs/1412.6980}
}

@inproceedings{selvaraju2017gradcam,
  author    = {Selvaraju, Ramprasaath R. and Cogswell, Michael and Das, Abhishek and Vedantam, Ramakrishna and Parikh, Devi and Batra, Dhruv},
  title     = {{Grad-CAM}: Visual Explanations from Deep Networks via Gradient-Based Localization},
  booktitle = ICCV,
  year      = {2017},
  url       = {https://openaccess.thecvf.com/content_iccv_2017/html/Selvaraju_Grad-CAM_Visual_Explanations_ICCV_2017_paper.html}
}

@inproceedings{cuturi2013sinkhorn,
  title     = {Sinkhorn Distances: Lightspeed Computation of Optimal Transport},
  author    = {Cuturi, Marco},
  booktitle = NeurIPS,
  year      = {2013},
  url       = {https://proceedings.neurips.cc/paper/2013/hash/af21d0c97db2e27e13572cbf59eb343d-Abstract.html}
}

@article{peyre2019computationalot,
  title   = {Computational Optimal Transport: With Applications to Data Science},
  author  = {Peyr{\'e}, Gabriel and Cuturi, Marco},
  journal = {Foundations and Trends in Machine Learning},
  volume  = {11},
  number  = {5--6},
  pages   = {355--607},
  year    = {2019},
  url     = {https://optimaltransport.github.io/pdf/ComputationalOT.pdf}
}

@article{banerjee2005vmf,
  title   = {Clustering on the Unit Hypersphere using von {M}ises-{F}isher Distributions},
  author  = {Banerjee, Arindam and Dhillon, Inderjit S. and Ghosh, Joydeep and Sra, Suvrit},
  journal = JMLR,
  volume  = {6},
  pages   = {1345--1382},
  year    = {2005},
  url     = {https://www.jmlr.org/papers/v6/banerjee05a.html}
}

@techreport{wah2011cub,
  title       = {The Caltech-UCSD Birds-200-2011 Dataset},
  author      = {Wah, Catherine and Branson, Steve and Welinder, Peter and Perona, Pietro and Belongie, Serge},
  institution = {California Institute of Technology},
  year        = {2011},
  number      = {CNS-TR-2011-001},
  url         = {https://www.vision.caltech.edu/datasets/cub_200_2011/}
}

@inproceedings{krause2013cars,
  author    = {Krause, Jonathan and Stark, Michael and Deng, Jia and Fei-Fei, Li},
  title     = {3D Object Representations for Fine-Grained Categorization},
  booktitle = ICCVW,
  year      = {2013},
  url       = {https://openaccess.thecvf.com/content_iccv_workshops_2013/W19/html/Krause_3D_Object_Representations_2013_ICCV_paper.html}
}

@inproceedings{KhoslaYaoJayadevaprakashFeiFei_FGVC2011,
author = "Aditya Khosla and Nityananda Jayadevaprakash and Bangpeng Yao and Li Fei-Fei",
title = "Novel Dataset for Fine-Grained Image Categorization",
booktitle = "First Workshop on Fine-Grained Visual Categorization, IEEE Conference on Computer Vision and Pattern Recognition",
year = "2011",
month = "June",
address = "Colorado Springs, CO",
url       = {http://vision.stanford.edu/aditya86/ImageNetDogs/}
}

@inproceedings{huang2023evalprotopnet,
  author    = {Huang, Qihan and Xue, Mengqi and Huang, Wenqi and Zhang, Haofei and Song, Jie and Jing, Yongcheng and Song, Mingli},
  title     = {Evaluation and Improvement of Interpretability for Self-Explainable Part-Prototype Networks},
  booktitle = ICCV,
  year      = {2023},
  pages     = {2011--2020},
  url       = {https://openaccess.thecvf.com/content/ICCV2023/html/Huang_Evaluation_and_Improvement_of_Interpretability_for_Self-Explainable_Part-Prototype_Networks_ICCV_2023_paper.html}
}

@inproceedings{xue2024protopformer,
  author    = {Xue, Mengqi and Huang, Qihan and Zhang, Haofei and Hu, Jingwen and Song, Jie and Song, Mingli and Jin, Canghong},
  title     = {{ProtoPFormer}: Concentrating on Prototypical Parts in Vision Transformers for Interpretable Image Recognition},
  booktitle = IJCAI,
  year      = {2024},
  pages     = {1516--1524},
  url       = {https://www.ijcai.org/proceedings/2024/168}
}

@article{oquab2024dinov2,
  author  = {Oquab, Maxime and Darcet, Timoth{\'e}e and Moutakanni, Th{\'e}o and Vo, Huy V. and Szafraniec, Marc and Khalidov, Vasil and Fernandez, Pierre and Haziza, Daniel and Massa, Francisco and El-Nouby, Alaa and Assran, Mahmoud and Ballas, Nicolas and Galuba, Wojciech and Howes, Russell and Huang, Po-Yao and Li, Shang-Wen and Misra, Ishan and Rabbat, Michael and Sharma, Vasu and Synnaeve, Gabriel and Xu, Hu and Jegou, Herv{\'e} and Labatut, Patrick and Joulin, Armand and Bojanowski, Piotr},
  title   = {{DINOv2}: Learning Robust Visual Features without Supervision},
  journal = TMLR,
  year    = {2024},
  url     = {https://arxiv.org/abs/2304.07193}
}

@inproceedings{yu2021ap10k,
  author    = {Yu, Hang and Xu, Yufei and Zhang, Jing and Zhao, Wei and Guan, Ziyu and Tao, Dacheng},
  title     = {{AP-10K}: A Benchmark for Animal Pose Estimation in the Wild},
  booktitle = {Proceedings of the Neural Information Processing Systems Track on Datasets and Benchmarks},
  year      = {2021},
  url       = {https://datasets-benchmarks-proceedings.neurips.cc/paper/2021/hash/903ce9225fca3e988c2af215d4e544d3-Abstract-round2.html}
}

@misc{carion2025sam3segmentconcepts,
  author       = {Nicolas Carion and Laura Gustafson and Yuan-Ting Hu and Shoubhik Debnath and Ronghang Hu and Didac Suris and Chaitanya Ryali and Kalyan Vasudev Alwala and Haitham Khedr and Andrew Huang and Jie Lei and Tengyu Ma and Baishan Guo and Arpit Kalla and Markus Marks and Joseph Greer and Meng Wang and Peize Sun and Roman R{\"a}dle and Triantafyllos Afouras and Effrosyni Mavroudi and Katherine Xu and Tsung-Han Wu and Yu Zhou and Liliane Momeni and Rishi Hazra and Shuangrui Ding and Sagar Vaze and Francois Porcher and Feng Li and Siyuan Li and Aishwarya Kamath and Ho Kei Cheng and Piotr Doll{\'a}r and Nikhila Ravi and Kate Saenko and Pengchuan Zhang and Christoph Feichtenhofer},
  title        = {{SAM 3}: Segment Anything with Concepts},
  howpublished = {arXiv:2511.16719},
  year         = {2025},
  url          = {https://arxiv.org/abs/2511.16719v2}
}
}
\maketitlesupplementary
\appendix
\section{Overview}
\label{sec:supp_overview}
We first describe the experimental setup and explain how we evaluate part explanations on CUB-200-2011 and AP-10K. We then detail how scoring and training change when classes share a prototype pool. The additional experiments report accuracy on Stanford Dogs and Stanford Cars, compare CNN and DINO backbones, and examine foreground gating and offline augmentation. We also report shared-pool consistency by prototype rank, accuracy after pruning, and training and inference costs. Finally, we derive the vMF overlap used in the diversity loss and present prediction-level why-tables and side-by-side comparisons of prototype activation maps.

\section{Experimental Details}
\label{sec:supp_setup}

\subsection{Datasets and Prototype Budgets}
We evaluate classification and part explanations on CUB-200-2011~\cite{wah2011cub} and AP-10K~\cite{yu2021ap10k}, and extend the classification evaluation to Stanford Dogs~\cite{KhoslaYaoJayadevaprakashFeiFei_FGVC2011} and Stanford Cars~\cite{krause2013cars}. For all four datasets, we train on the full official training split. Class-specific models use five or seven prototypes per class, whereas for shared models we match the total number of foreground prototypes across methods. These totals are 512 or 1,000 on CUB and 125 or 250 on AP-10K. The main comparisons use DINOv2 ViT-B/14~\cite{oquab2024dinov2} and DINOv3 ViT-B/16~\cite{simeoni2025dinov3}, and on CUB we also evaluate the class-specific models with the corresponding Small backbones.

\subsection{Foreground Extraction}
\label{subsec:supp_foreground_extraction}

We first use attention to narrow the set of patches that PCA will consider. For an image $x$, let $A^{(L,h)}_{\mathrm{cls}\to i}(x)$ be the attention from the class token to patch $i$ in head $h$ of frozen block $L$. Averaging over its $H$ heads gives one attention score per patch:
\begin{equation}
m_i^{\mathrm{attn}}(x)=\frac{1}{H}\sum_{h=1}^{H} A^{(L,h)}_{\mathrm{cls}\to i}(x).
\label{eq:attn_gating}
\end{equation}
We smooth and normalize the resulting attention map, then threshold it to obtain the candidate region $\mathcal{R}(x)$.

Within $\mathcal{R}(x)$, we refine the attention mask using the variation in the frozen patch tokens. We normalize each frozen token to unit length, giving $\boldsymbol{u}_i$, and center the tokens by subtracting their mean $\bar{\boldsymbol{u}}_{\mathcal{R}}$ within the region. We then compute the first principal direction $\boldsymbol{v}$ from the centered tokens and project each candidate patch $i\in\mathcal{R}(x)$ onto it:
\begin{equation}
s_i(x)=\boldsymbol{v}^{\top}\bigl(\boldsymbol{u}_i-\bar{\boldsymbol{u}}_{\mathcal{R}}\bigr).
\label{eq:refine_pca}
\end{equation}

The projections order the patches along the principal axis, but the arbitrary sign of $\boldsymbol{v}$ leaves open which end corresponds to the foreground. After rescaling the projections within $\mathcal{R}(x)$ to $[0,1]$, we split the patches at the median projection and compare their mean attention on either side. If the group with higher projections has lower mean attention, we reverse the scores. Thresholding the oriented scores $m_i(x)$ at $\tau_{\mathrm{PCA}}$ gives the binary mask $w(x)$, with $w_i(x)=0$ outside $\mathcal{R}(x)$. If there are too few patches for refinement or PCA cannot separate the candidate region, we keep the attention mask. \Cref{tab:supp_hyperparameters} lists the attention and PCA thresholds used on CUB and AP-10K.

\subsection{Discovery and Fitting}
We train for 25 epochs with Adam~\cite{kingma2015adam}. In the CUB reference setting, we use the first epoch to initialize the prototype directions with optimal transport, then continue discovery through epoch 12. We introduce the fitted parameters in epoch 13 and use the fully fitted model from epoch 14 through epoch 25. To fit each prototype, we weight its assigned patches by their cosine alignment with the discovered direction and normalize the weights within each image so that every supporting image contributes the same total. The query batch, on which we compute the current loss, is excluded from that fit. As the fitted parameters take over, we retain prediction distillation and reference-peak supervision while phasing out PPD. Once training is complete, we refresh the assignments on the training set and refit the prototypes with the features and class weights held fixed.

\section{Explanation Metrics}
\label{sec:supp_metrics}
\begin{table*}[t]
  \centering
  \small
  \caption{\textbf{Implementation settings for the class-specific \method{} model.} We report the training schedule, optimization settings, and foreground thresholds.}
  \label{tab:supp_hyperparameters}
  \begin{tabular}{@{}ll@{}}
    \toprule
    \textbf{Setting} & \textbf{Value} \\
    \midrule
    Input & $224\times224$\\
    Warm-up, discovery, transition, fully fitted & Epochs 1, 2--12, 13, and 14--25, respectively \\
    OT warm-up & Momentum $0.99$, entropy coefficient $0.05$, 10 Sinkhorn iterations, initial $\kappa=1$ \\
    Added block learning rate & $10^{-4}$ in both stages \\
    Mean direction and concentration learning rates & $5\times10^{-4}$ and $7\times10^{-4}$ during discovery \\
    Class weight learning rate & $10^{-5}$ in discovery and $10^{-3}$ during fitting \\
    Discovery losses & PPD weight $0.8$, LogDet weight $1$ \\
    Prediction distillation and reference peak & KL weight $0.5$ for six adaptation epochs, then decays to $0.1$. Reference-peak weight $4$ \\
    Foreground thresholds & Attention threshold $0.01$, PCA threshold $0.50$ \\
    \bottomrule
  \end{tabular}
\end{table*}
\subsection{CUB-200-2011}
We center a $72$-pixel window on each prototype's activation peak in the resized $224\times224$ image and record the visible part landmarks inside it. Using these matches, we follow \citet{huang2023evalprotopnet} and consider a prototype consistent for a class if there is a part that it matches in at least $80\%$ of the images where that part is visible. To assess the stability of these matches, we repeat the peak search after adding Gaussian noise to the normalized input, with standard deviation $0.2$ and clipping at $\pm0.25$. We then compare the sets of parts matched in the clean and perturbed views, counting a response as stable when the sets are identical, including when both are empty. We average both metrics over class--prototype pairs.

While consistency and stability assess the reliability of a prototype's matches, distinctiveness measures whether different prototypes supply evidence from different locations. To quantify this separation, we compute the mean pairwise intersection over union of their peak-centered windows within each image and report its complement as a percentage~\cite{zhu2025nonparampartprototypes}. For shared models, we evaluate the five highest-weighted prototypes for each class, keeping that selection fixed across images of the class.

\subsection{AP-10K}
On AP-10K, the original landmarks locate points on the legs and face but do not delineate regions such as the body and tail. We therefore evaluate the head, nose, eyes, legs, body, and tail as six part groups. Within the leg group, we combine the visible shoulder, elbow, front-paw, hip, knee, and back-paw landmarks without distinguishing left from right or front from hind legs. For the remaining groups, we use regions derived with SAM~3.1~\cite{carion2025sam3segmentconcepts}, merging the eyes across sides and including the ears and horns in the head group. With these groups defined, we evaluate consistency, stability, and distinctiveness analogously to the CUB protocol.

\section{Shared Prototype Pool}
\label{subsec:shared_pool}

We replace the separate prototype banks of \S~3.3 with a single pool of $R$ vMF prototypes. The foreground gate and normalized patch features stay the same. Since each prototype has one direction and concentration for all classes, we compute its log-density score $a_{i,j}=\log C_D(\kappa_j)+\kappa_j\boldsymbol{\mu}_j^\top\hat{\boldsymbol{t}}_i$ once per patch. Instead of choosing a winner for each class, we use a soft assignment over the shared pool, $q_{i,j}=\operatorname{softmax}_j(a_{i,j}/\tau_q)$, where $\tau_q$ is the assignment temperature. Each class then applies its own mixture weights $\pi_{c,j}\geq 0$, with $\sum_{j=1}^{R}\pi_{c,j}=1$. For a nonempty foreground set $\mathcal F(x)$, the patch evidence and image logit are
\begin{equation}
e_{i,c}=\sum_{j=1}^{R}q_{i,j}\pi_{c,j}a_{i,j},
\qquad
\ell_c(x)=\frac{1}{|\mathcal F(x)|}\sum_{i\in\mathcal F(x)}e_{i,c}.
\end{equation}
The class thus changes how much a match contributes, not how the match is computed. As in \S~3.3, a softmax over the image logits gives the class probabilities.

Sharing also changes which patches compete during discovery. We initialize the pool with optimal transport within each class, then solve a single foreground transport problem across the classes in each batch. A prototype's target mass is the average of its mixture weights for the classes of the images in that batch, weighted by each image's foreground patch count. The hard assignments from this plan provide the targets for patch-to-prototype distillation. LogDet still acts within each class, but now compares the five shared prototypes with the highest mixture weights for that class.

For statistical fitting, each foreground patch still goes to a single prototype. We select the highest-scoring prototype in the entire pool rather than in the image's labeled class. We keep the within-image patch weights of \S~3.4 and scale the total contribution of an image with label $y$ to prototype $j$ by $R\pi^0_{y,j}$, where $\pi^0$ is the mixture-weight matrix at the end of discovery. The label therefore affects how much the patch contributes to the fit, not which prototype receives it. The weighted moment fit, transition to fitted parameters, adaptation losses, and final refit otherwise follow \S~3.4.

\section{Additional Quantitative Results}
\label{sec:supp_quant}

\begin{table*}[t]
  \centering
  \small
  \caption{\textbf{Stanford Dogs and Stanford Cars top-1 accuracy (\%).} $J$ is the number of prototypes per class. The highest value in each column is bold, including ties at the displayed precision.}
  \label{tab:dogs_cars_supp}
  \begin{tabular}{lc|cc|cc}
    \toprule
    & & \multicolumn{2}{c|}{\textbf{Stanford Dogs}} & \multicolumn{2}{c}{\textbf{Stanford Cars}} \\
    \cmidrule(lr){3-4}\cmidrule(lr){5-6}
    Method & $J$ & DINOv2 ViT-B/14 & DINOv3 ViT-B/16 & DINOv2 ViT-B/14 & DINOv3 ViT-B/16 \\
    \midrule
    
ProtoPNet      & 5 & 44.1 & 82.1 &  49.9 & 79.0 \\
ProtoPNet      & 7 & 49.2 & 81.8 & 50.0 & 85.9 \\
\midrule
Def.\ ProtoPNet & 5 & 78.6 & 81.4 & 91.2 & 90.0 \\
Def.\ ProtoPNet & 7 & 80.8 & 83.2 & \textbf{91.4} & 92.4 \\
\midrule
TesNet         & 5 & 80.5 & 85.3 & 90.7 & 85.1 \\
TesNet         & 7 & 79.6 & 85.5 & 89.4 & 92.6 \\
\midrule
EvalProtoPNet  & 5 & 80.6 & 81.0 & 78.7 & 82.9 \\
EvalProtoPNet  & 7 & 79.3 & 81.9 & 81.4 & 84.7 \\
\midrule
MGProto        & 5 & 76.4 & 82.4 & 85.2 & 93.5 \\
MGProto        & 7 & 74.7 & 83.0 &  84.6 & 92.9 \\
\midrule
NPPP           & 5 & 88.0 & 86.9 & 90.1 & 80.5 \\
NPPP           & 7 & \textbf{88.2} & 87.0 & 90.8 & 86.3 \\

    \midrule
    \method{}     & 5 & 87.8 & 87.5 & \textbf{91.4} & 93.2 \\
    \method{}     & 7 & 87.6 & \textbf{87.8} & \textbf{91.4} & \textbf{93.8} \\
    \bottomrule
  \end{tabular}
\end{table*}

\begin{table*}[t]
  \centering
  \small
  \caption{\textbf{Published CNN references and DINOv2 runs on CUB.} Accuracies are in percent. $J$ counts prototypes or basis vectors per class. Bold marks the higher reported accuracy in each row.}
  \label{tab:supp_cnn_dino}
  \begin{tabular}{@{}llcc@{}}
    \toprule
    Method & Published CNN reference & CNN Acc., $J=10$ & DINOv2 ViT-B/14, $J=7$ \\
    \midrule
    Deformable ProtoPNet & ResNet-50, full images~\cite[Table~3 (top)]{donnelly2022deformableprotopnet} & 86.4 & \textbf{87.9} \\
    TesNet & DenseNet-121, cropped images~\cite[Table~1 (top)]{wang2021tesnet} & $84.8\pm0.2$ & \textbf{86.8} \\
    \bottomrule
  \end{tabular}
\end{table*}
\subsection{Cross-Dataset Accuracy}
\label{subsec:supp_cross_dataset}
We extend the classification evaluation to Stanford Dogs and Stanford Cars by training a separate classifier on each official training split and reporting top-1 accuracy. With the two Base backbones and five or seven prototypes per class, our models reach $87.5$--$87.8\%$ on Dogs, with little effect from changing the prototype budget. On Cars, the choice of backbone makes a larger difference, with accuracy reaching $91.4\%$ with DINOv2 and $93.2$--$93.8\%$ with DINOv3.

\subsection{CNN and DINO Backbones}
All methods in the main CUB comparisons use DINO, so we also put the baseline performance in context by comparing Deformable ProtoPNet~\cite{donnelly2022deformableprotopnet} and TesNet~\cite{wang2021tesnet} with their published single-model CNN references. For this comparison, we use our DINOv2 ViT-B/14 runs with $J=7$, which is closer to the published CNN budget of $J=10$ than our $J=5$ setting. Both methods achieve higher accuracy in these DINO runs (\cref{tab:supp_cnn_dino}).

\subsection{Additional Ablations}
\label{sec:supp_ablations}

\subsubsection{Foreground Gating}
\label{subsec:supp_fg_ablation}
We next examine whether foreground gating accounts for the explanation gains on CUB by giving NPPP and ProtoPNet the same attention and PCA gate used by \method{}. In DINOv2 ViT-B/14 controls with $J=5$, this improves accuracy and consistency for both baselines. Even with foreground handling matched in this way, fitted \method{} leads both on consistency, stability, and distinctiveness, while NPPP retains a small accuracy lead (\cref{tab:supp_matched_gate}).

We also vary the gate within \method{}, where attention provides PCA with a reference for resolving the foreground direction. Combining the two steps gives the highest consistency, stability, and accuracy (\cref{tab:fg_ablation_supp}), whereas neither step alone matches the full gate on these metrics. Removing the gate altogether reduces consistency further, producing the largest drop. This may reflect the diversity pressure from LogDet, which penalizes overlap between prototype distributions. Without foreground gating, some prototypes can satisfy this pressure by matching background patterns rather than object parts.

\begin{table}[t]
  \centering
  \small
  \setlength{\tabcolsep}{3.3pt}
  \caption{\textbf{Matched foreground gate on CUB.} DINOv2 ViT-B/14, $J=5$. All values are percentages. The highest reported value in each column is bold.}
  \label{tab:supp_matched_gate}
  \begin{tabular}{@{}llrrrr@{}}
    \toprule
    Method & Gate & Con. & Sta. & Dis. & Acc. \\
    \midrule
    ProtoPNet & Native & 8.0 & 62.6 & 90.3 & 64.3 \\
    ProtoPNet & Ours & 10.3 & 55.2 & 73.8 & 68.6 \\
    NPPP & Native & 58.5 & 81.0 & 81.0 & 90.8 \\
    NPPP & Ours & 61.3 & 81.5 & 85.1 & \textbf{91.1} \\
    \method{} & Ours & \textbf{80.2} & \textbf{84.3} & \textbf{95.9} & 90.5 \\
    \bottomrule
  \end{tabular}
\end{table}

\begin{table}[t]
  \centering
  \small
  \setlength{\tabcolsep}{3.3pt}
  \caption{\textbf{Foreground gating ablation for \method{} on CUB.} DINOv2 ViT-B/14, $J=5$. All values are percentages, and the highest value for each metric is bold.}
  \label{tab:fg_ablation_supp}
  \begin{tabular}{@{}lrrrr@{}}
    \toprule
    Foreground gating & Con. & Sta. & Dis. & Acc. \\
    \midrule
    Attention then PCA (default) & \textbf{80.2} & \textbf{84.3} & 95.9 & \textbf{90.5} \\
    PCA only       & 56.0 & 75.8 & 95.7 & 88.7 \\
    Attention only & 66.6 & 79.4 & 96.7 & 90.1 \\
    None           & 22.8 & 75.9 & \textbf{96.8} & 89.5 \\
    \bottomrule
  \end{tabular}
\end{table}

\subsubsection{Offline Augmentation}
\label{subsec:supp_offline_aug}
The main comparison uses no offline augmentation, so we examine its effect in a separate CUB experiment with DINOv2 ViT-B/14 and $J=5$. We compare training without offline augmentation to a ProtoPNet-style recipe that generates 40 variants per training image by rotation, skew, shear, and horizontal flipping, then trains for five epochs on the expanded set~\cite{chen2019thislookslikethat}. Under this recipe, NPPP and EvalProtoPNet improve on all four metrics, while the other baselines show mixed changes. In contrast, the scores of \method{} change little, and even without offline augmentation it outscores every augmented baseline on all three explanation metrics (\cref{tab:offline_aug_all_supp}).

\begin{table}[t]
  \centering
  \small
  \setlength{\tabcolsep}{3.1pt}
  \renewcommand{\arraystretch}{1.02}
  \caption{\textbf{Effect of offline data augmentation across methods.} Con., Sta., Dis., and Acc. are reported in \% for CUB-200-2011 (DINOv2 ViT-B/14, $J{=}5$). The highest reported value in each column is bold.}
  \label{tab:offline_aug_all_supp}
  \begin{tabular}{@{}llrrrr@{}}
    \toprule
    Method & Offline aug. & Con. & Sta. & Dis. & Acc. \\
    \midrule 
ProtoPNet & No & 8.0 & 62.6 & 90.3 & 64.3 \\
ProtoPNet & Yes & 19.6 & 48.2 & 66.5 & 80.8 \\
\midrule
Def. ProtoPNet & No & 18.4 & 73.4 & 65.8 & 86.7 \\
Def. ProtoPNet & Yes & 51.1 & 77.4 & 34.4 & 76.4 \\
\midrule
TesNet & No & 28.6 & 74.8 & 37.8 & 85.2 \\
TesNet & Yes & 57.3 & 69.9 & 54.8 & 86.1 \\
\midrule
EvalProtoPNet & No & 53.8 & 71.2 & 36.5 & 83.2 \\
EvalProtoPNet & Yes & 65.6 & 73.2 & 43.1 & 87.5 \\
\midrule
MGProto & No & 16.7 & 68.9 & 53.2 & 83.7 \\
MGProto & Yes & 29.7 & 72.4 & 39.1 & 78.4 \\
\midrule
NPPP & No & 58.5 & 81.0 & 81.0 & 90.8 \\
NPPP & Yes & 69.7 & 82.0 & 95.2 & \textbf{91.1} \\
    \midrule
    \method{} & No & \textbf{80.2} & \textbf{84.3} & 95.9 & 90.5 \\
    \method{} & Yes & 79.0 & 83.6 & \textbf{96.3} & 90.3 \\
    \bottomrule
  \end{tabular}
\end{table}

\subsection{Shared-Pool Consistency and Pruning}
\label{subsec:supp_pruning}
\begin{figure*}[!t]
  \centering
  \includegraphics[trim=302bp 0 0 0,clip,width=0.29\linewidth]{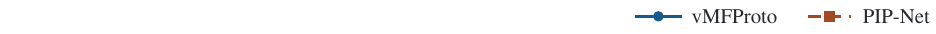}\par
  \begin{subfigure}[t]{0.48\linewidth}
    \vspace{0pt}
    \centering
    \includegraphics[width=\linewidth]{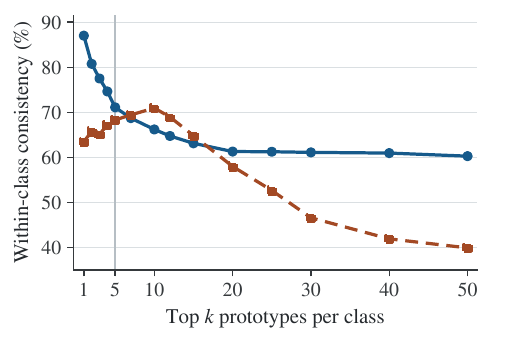}
    \caption{\textbf{Consistency by rank.}}
    \label{fig:shared_topk_consistency}
  \end{subfigure}\hspace{0.02\linewidth}%
  \begin{subfigure}[t]{0.48\linewidth}
    \vspace{0pt}
    \centering
    \includegraphics[width=\linewidth]{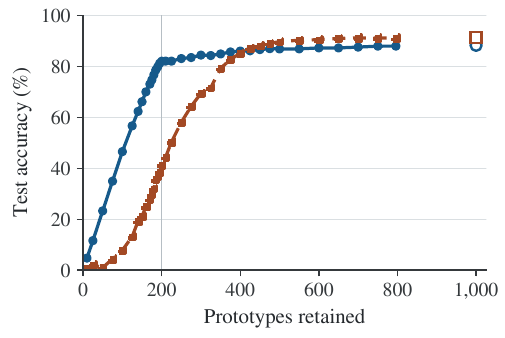}
    \caption{\textbf{Accuracy after pruning.}}
    \label{fig:shared_pruning_accuracy}
  \end{subfigure}
  \caption{\textbf{Shared-pool consistency and pruning on CUB.} (a) Consistency of each class's top $k$ prototypes by weight, averaged across classes. At large $k$, PIP-Net includes zero-weight prototypes. (b) Test accuracy at matched pool sizes. Open markers show unpruned pools.}
  \label{fig:shared_topk_pruning}
\end{figure*}

For shared pools, shortening an explanation leaves the classifier intact, whereas pruning changes the pool used for prediction. We compare these choices on CUB using DINOv2 ViT-B/14 and an initial pool of 1,000 prototypes for both \method{} and PIP-Net (\cref{fig:shared_topk_pruning}). For each value of $k$, we evaluate consistency using only each class's $k$ highest-weighted prototypes and average across classes (Fig.~\ref{fig:shared_topk_pruning}\subref{fig:shared_topk_consistency}). The reported consistency therefore depends on $k$, because changing $k$ changes the set of prototypes included in the average while leaving the classifier unchanged. Under this evaluation, \method{} has its largest consistency lead at $k=1$, although the ranking of the two models changes as more prototypes are shown.

To reduce the classifier itself, we rank each prototype by its best rank among classes where it has positive weight, breaking ties by its share of the weight in that class. We then prune both pools to matched sizes without retraining (Fig.~\ref{fig:shared_topk_pruning}\subref{fig:shared_pruning_accuracy}). PIP-Net starts with higher accuracy, but \method{} overtakes it under heavy pruning. The distinction between explanation length and classifier size also appears when we retain only the 742 \method{} prototypes that occur in at least one class's top five. Although all three top-five explanation scores remain unchanged, accuracy falls from $88.23\%$ to $87.52\%$, showing that prototypes outside every class's top five can still contribute to classification.

\subsection{Computational Cost}
\label{subsec:supp_runtime}
Finally, \cref{tab:supp_runtime} compares training time over 25 epochs and inference throughput on CUB with DINOv2 ViT-B/14 and $J=5$. We benchmark inference on the same NVIDIA RTX 6000 Ada GPU with batches of 32, excluding data loading, transfers, and explanation evaluation. Neither model applies a foreground gate in the timed forward pass, which for \method{} means bypassing the gate only at inference on an already trained checkpoint. This bypass leaves its accuracy nearly unchanged and its explanation metrics unchanged. Under this timing protocol, \method{} takes longer to train and reaches $78\%$ of NPPP's inference throughput.

\begin{table}[t]
  \centering
  \small
  \caption{\textbf{Training and inference cost on CUB.} DINOv2 ViT-B/14 with $J=5$. Training times cover 25 epochs.}
  \label{tab:supp_runtime}
  \begin{tabular}{@{}lcc@{}}
    \toprule
    \textbf{Method} & \textbf{Training (min)} & \textbf{Inference (img/s)} \\
    \midrule
    NPPP & \textbf{20.0} & \textbf{361.8} \\
    \method{} & 50.9 & 281.0 \\
    \bottomrule
  \end{tabular}
\end{table}

\FloatBarrier

\section{Additional Derivations}
\label{sec:supp_derivations}
\subsection{Closed-Form vMF Overlap}
\label{subsec:supp_vmf_overlap_derivation}
The LogDet penalty compares prototypes through the normalized $L^2$ overlap of their densities. To compute this overlap, we first multiply $f_{c,j}$ and $f_{c,k}$, which adds their exponent vectors. The exponential factor in this product therefore has the form of a vMF density with concentration
\begin{equation}
r_{c,jk}
=\left\|\kappa_{c,j}\boldsymbol\mu_{c,j}
+\kappa_{c,k}\boldsymbol\mu_{c,k}\right\|_2.
\end{equation}
Recognizing this vMF form gives the integral of the exponential over the unit sphere as $1/C_D(r_{c,jk})$. Restoring the two original normalizers then gives the overlap
\begin{equation}
S^{(c)}_{jk}
=\langle f_{c,j},f_{c,k}\rangle_{L^2}
=\frac{C_D(\kappa_{c,j})C_D(\kappa_{c,k})}{C_D(r_{c,jk})}.
\end{equation}
To normalize the overlap, we pair each prototype with itself, giving $r_{c,jj}=2\kappa_{c,j}$, and use the same expression to obtain its self-overlap. Dividing $S^{(c)}_{jk}$ by the geometric mean of the two self-overlaps then gives the kernel in the main text
\begin{equation}
K^{(c)}_{jk}
=\frac{S^{(c)}_{jk}}{\sqrt{S^{(c)}_{jj}S^{(c)}_{kk}}}
=\frac{\sqrt{C_D(2\kappa_{c,j})C_D(2\kappa_{c,k})}}{C_D(r_{c,jk})}.
\end{equation}

\section{Additional Qualitative Results}
\label{sec:supp_qual}
We first inspect which prototypes contribute to individual predictions, then compare activation maps across methods on the same test images to see where those prototypes respond.

\subsection{Additional Activation-Reasoning Why-Tables}
\label{subsec:supp_why_tables}
Each why-table shows the four prototypes that contribute the most positive evidence to a test prediction. For each of these prototypes, we pair its exemplar patch and training image with its activation map on the test image, and report its share of all positive evidence for the predicted class. On the test image, the white box marks the patch that contributed most for that prototype.

We show six selected predictions per panel, covering CUB-200-2011, AP-10K, Stanford Dogs, and Stanford Cars for the class-specific models (\cref{tab:supp_why_cub,tab:supp_why_ap10k,tab:supp_why_dogs,tab:supp_why_cars}) and CUB-200-2011 and AP-10K for the shared-pool models (\cref{tab:supp_why_cub_shared,tab:supp_why_ap10k_shared}). For these shared-pool panels, we search the full training split without filtering by class and label each training image with its source class. The Ringed Kingfisher example illustrates this cross-class search, with the strongest supporting prototype represented by a Green Kingfisher exemplar.

\begin{table*}[p]
  \centering
  \includegraphics[width=\textwidth,height=0.95\textheight,keepaspectratio]{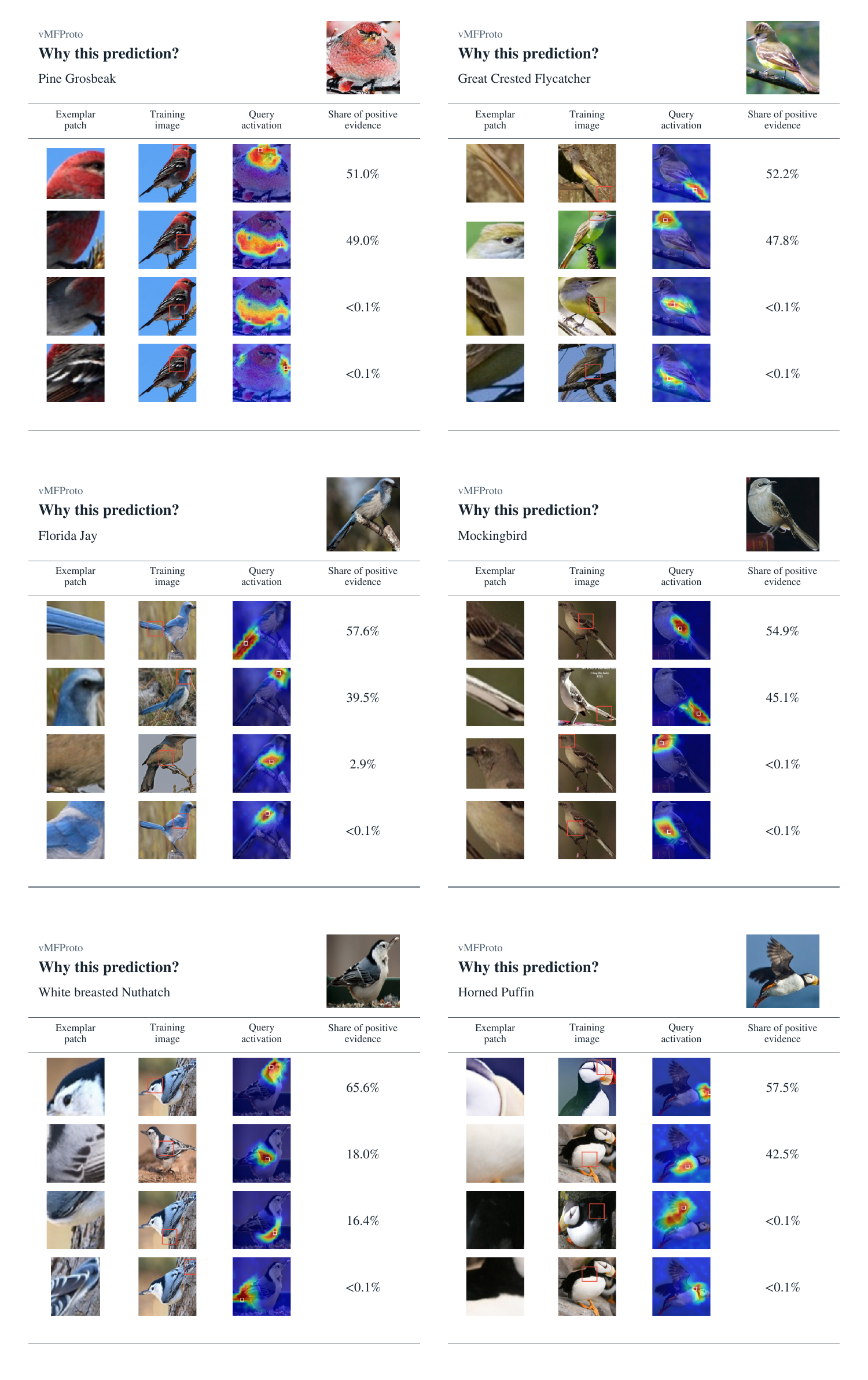}
  \caption{\textbf{Class-specific why-tables on CUB-200-2011.} Prototype evidence for six selected test predictions.}
  \label{tab:supp_why_cub}
\end{table*}

\begin{table*}[p]
  \centering
  \includegraphics[width=\textwidth,height=0.95\textheight,keepaspectratio]{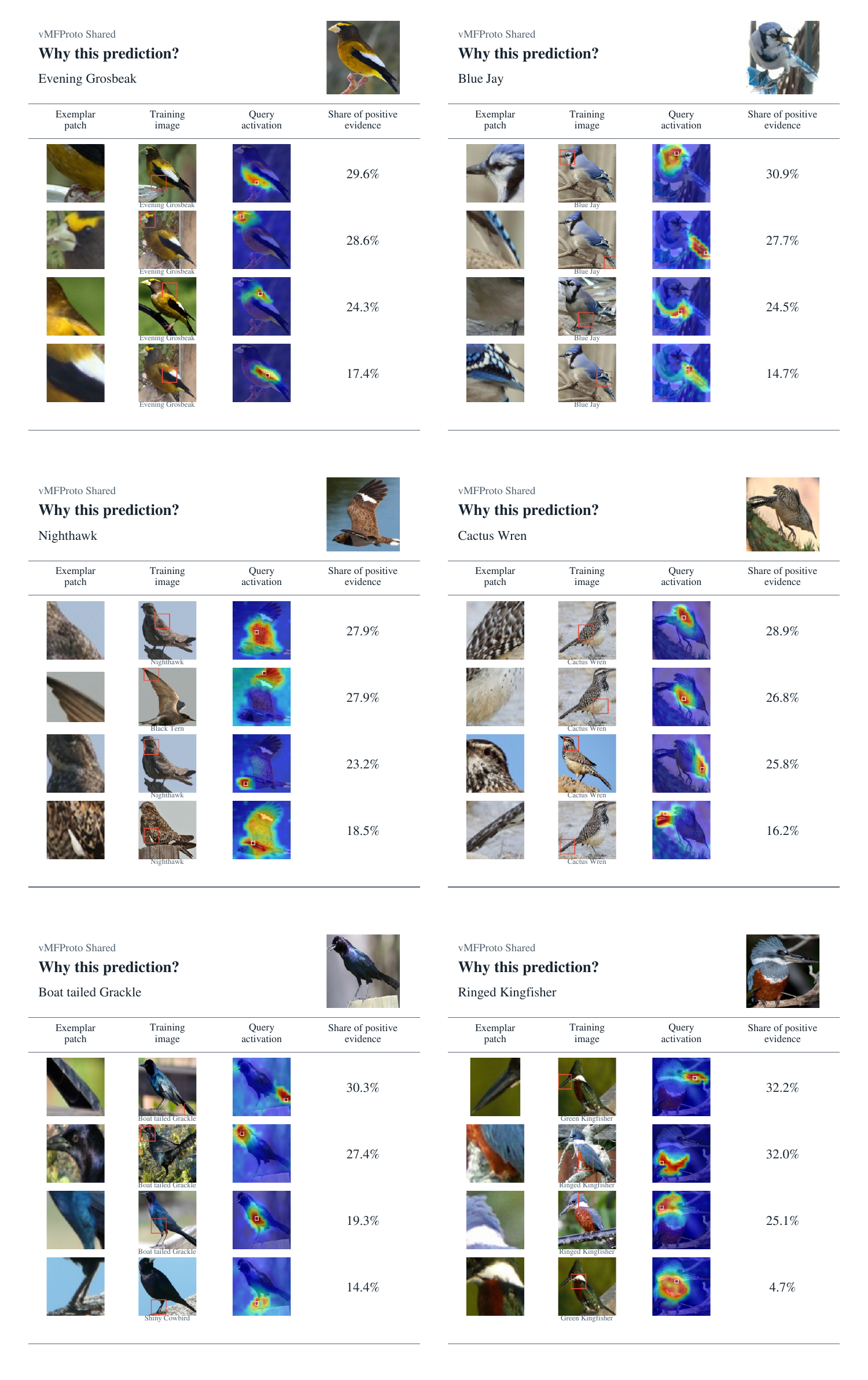}
  \caption{\textbf{Shared-pool why-tables on CUB-200-2011.} Exemplar classes are shown below the training images.}
  \label{tab:supp_why_cub_shared}
\end{table*}

\begin{table*}[p]
  \centering
  \includegraphics[width=\textwidth,height=0.95\textheight,keepaspectratio]{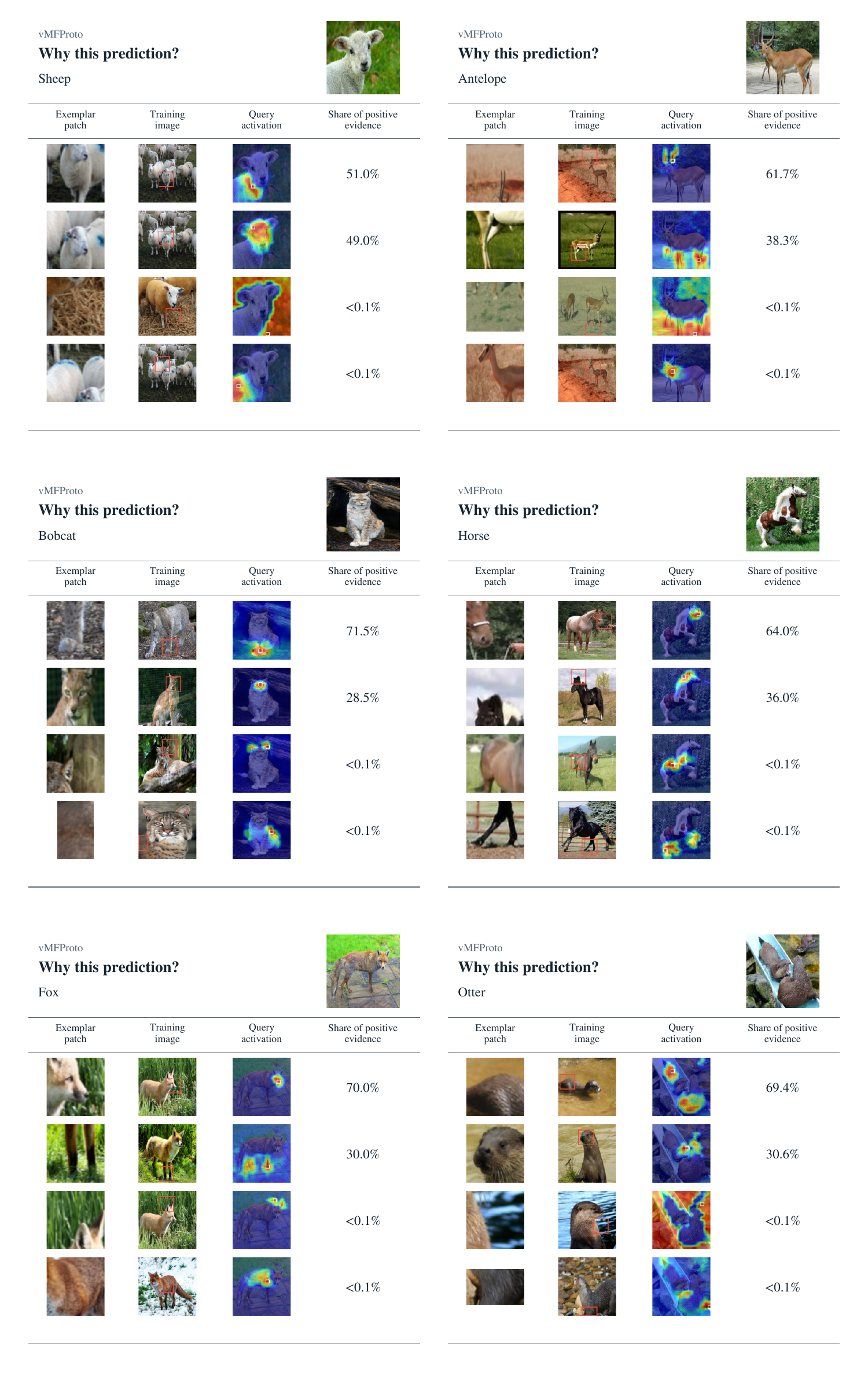}
  \caption{\textbf{Class-specific why-tables on AP-10K.} Prototype evidence for six selected test predictions.}
  \label{tab:supp_why_ap10k}
\end{table*}

\begin{table*}[p]
  \centering
  \includegraphics[width=\textwidth,height=0.95\textheight,keepaspectratio]{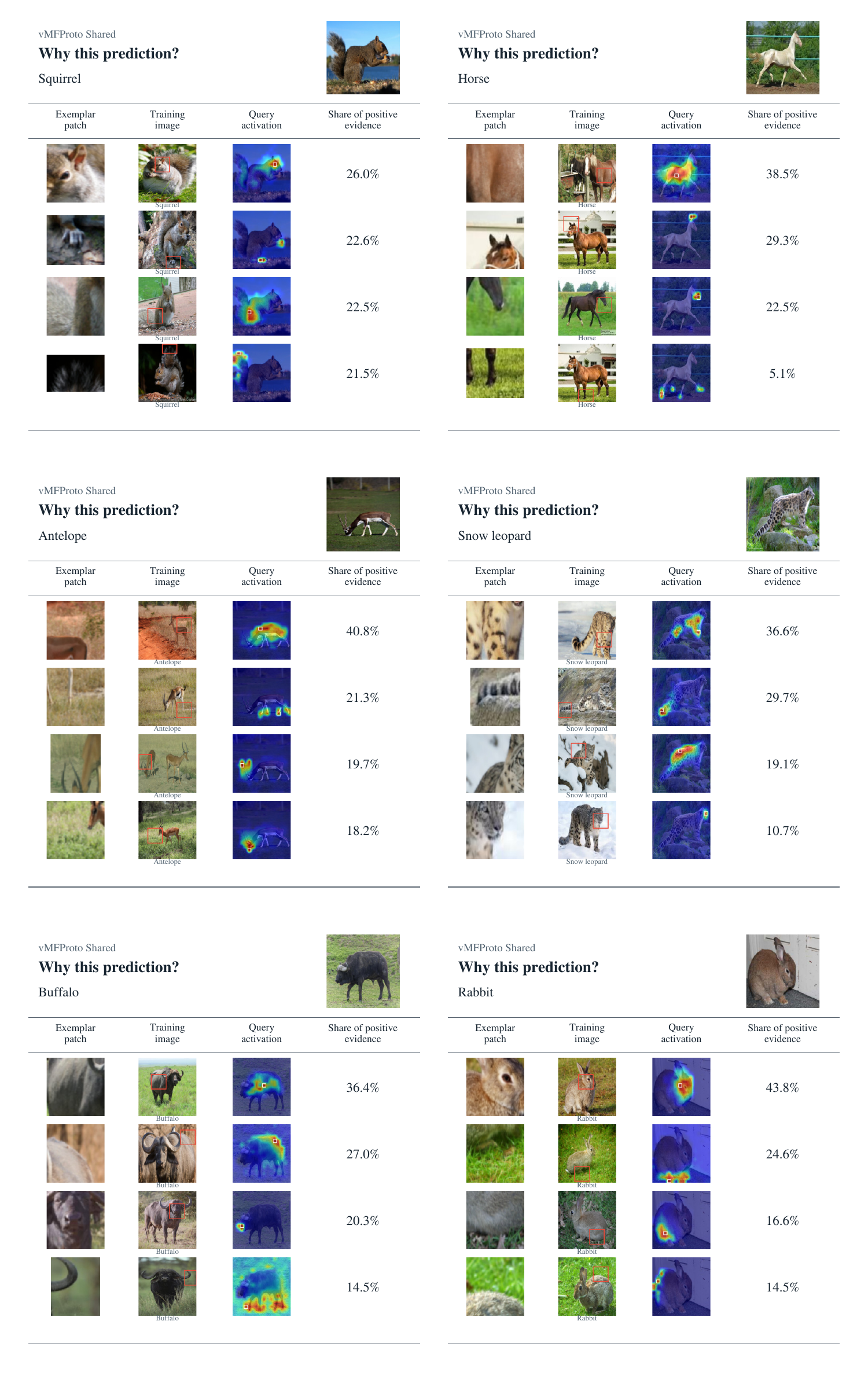}
  \caption{\textbf{Shared-pool why-tables on AP-10K.} Exemplar classes are shown below the training images.}
  \label{tab:supp_why_ap10k_shared}
\end{table*}

\begin{table*}[p]
  \centering
  \includegraphics[width=\textwidth,height=0.95\textheight,keepaspectratio]{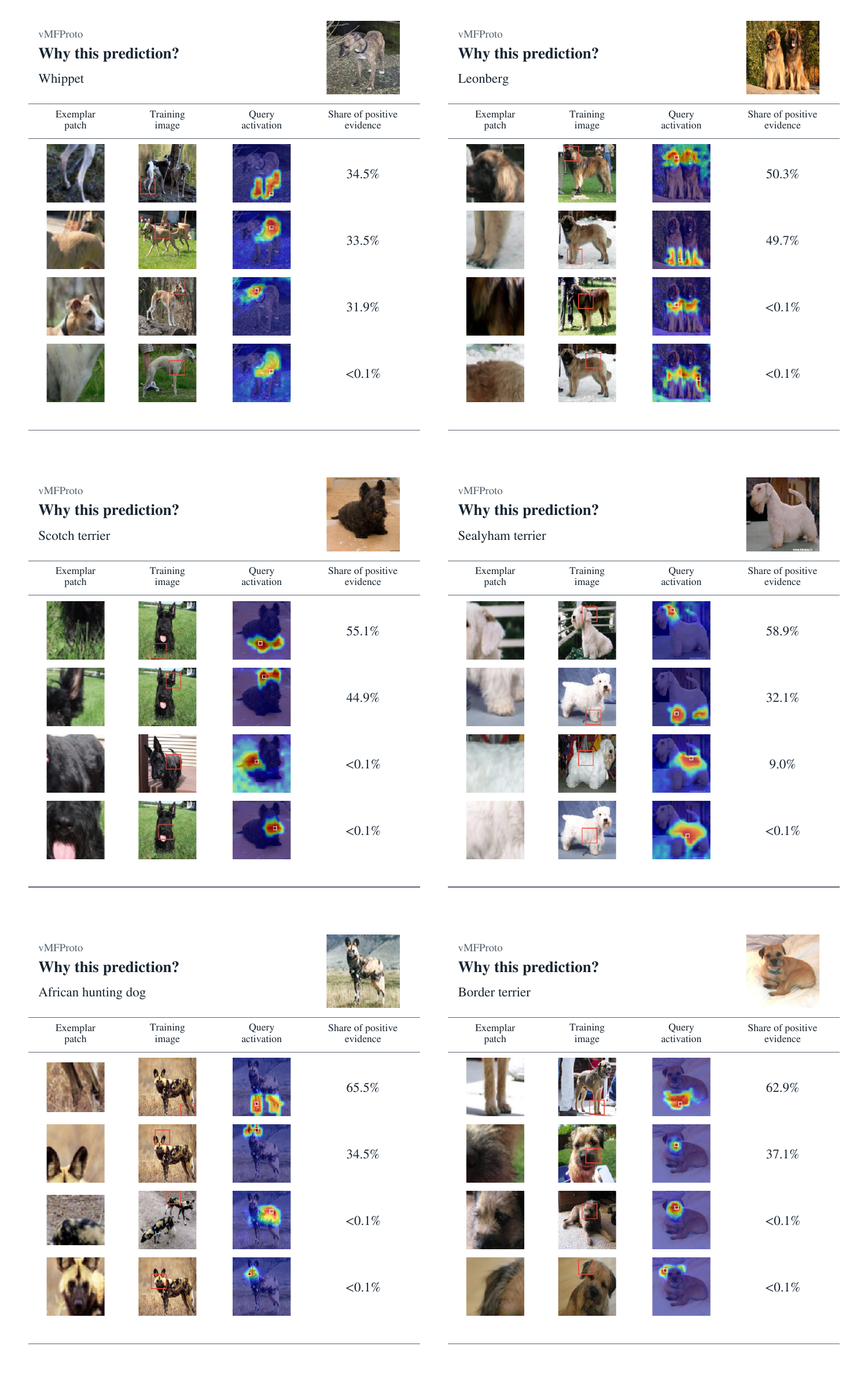}
  \caption{\textbf{Class-specific why-tables on Stanford Dogs.} Prototype evidence for six selected test predictions.}
  \label{tab:supp_why_dogs}
\end{table*}

\begin{table*}[p]
  \centering
  \includegraphics[width=\textwidth,height=0.95\textheight,keepaspectratio]{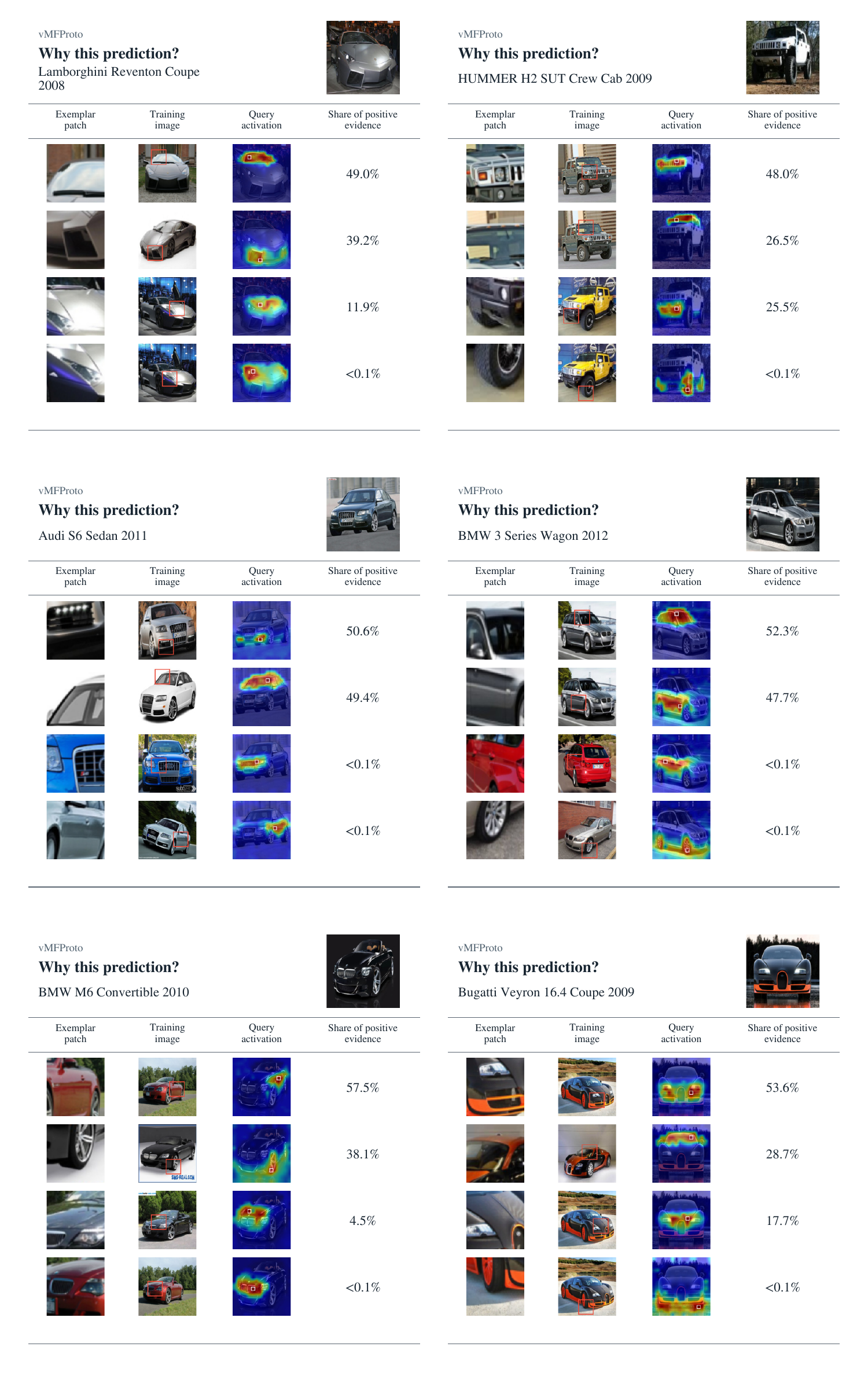}
  \caption{\textbf{Class-specific why-tables on Stanford Cars.} Prototype evidence for six selected test predictions.}
  \label{tab:supp_why_cars}
\end{table*}

\subsection{Additional Baseline Qualitative Comparisons}
\begin{figure*}[t]
  \centering
  \includegraphics[width=\textwidth,height=0.83\textheight,keepaspectratio]{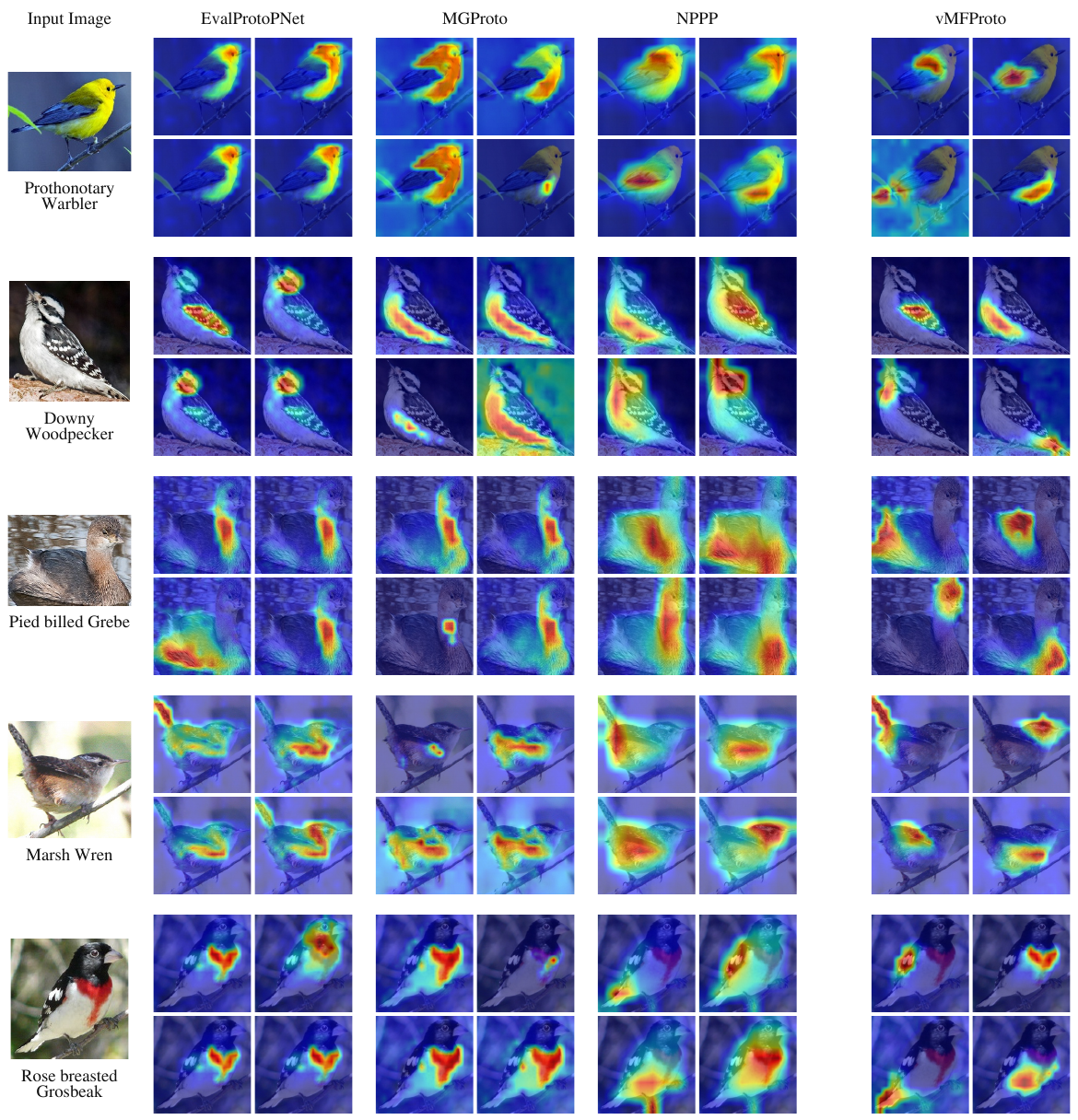}
  \caption{\textbf{Class-specific prototypes on CUB-200-2011.} Each row shows the input image and the top four prototype activation maps for its ground-truth class. All methods use DINOv2 ViT-B/14 and $J{=}5$ prototypes per class.}
  \label{fig:supp_cub_qual}
\end{figure*}

\begin{figure*}[t]
  \centering
  \includegraphics[width=\textwidth,height=0.83\textheight,keepaspectratio]{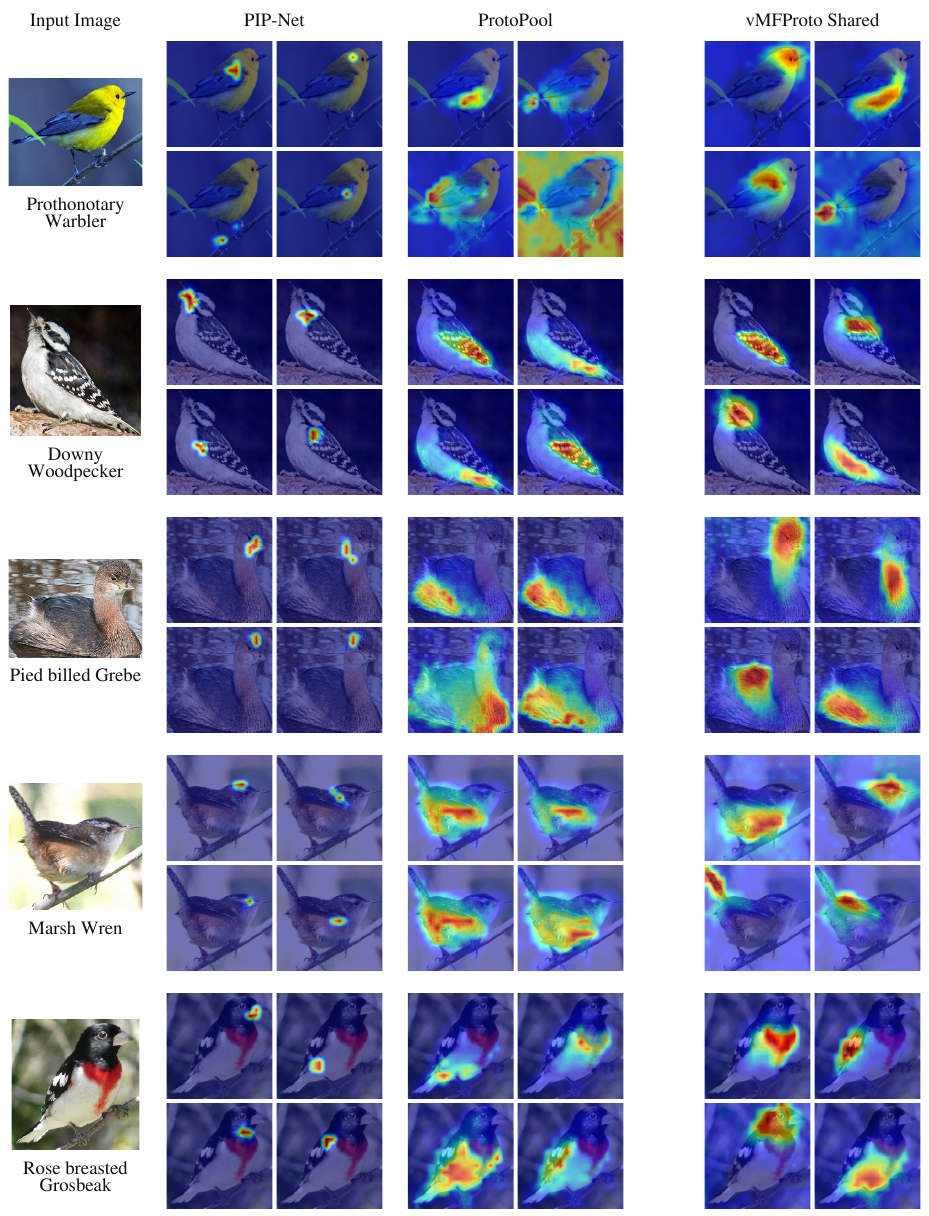}
  \caption{\textbf{Shared prototypes on CUB-200-2011.} For the same images as in \cref{fig:supp_cub_qual}, we show each method's four highest-ranked shared prototypes for the ground-truth class. All methods use DINOv2 ViT-B/14 and a pool of $R{=}1{,}000$ prototypes.}
  \label{fig:supp_cub_shared_qual}
\end{figure*}

\begin{figure*}[t]
  \centering
  \includegraphics[width=\textwidth,height=0.83\textheight,keepaspectratio]{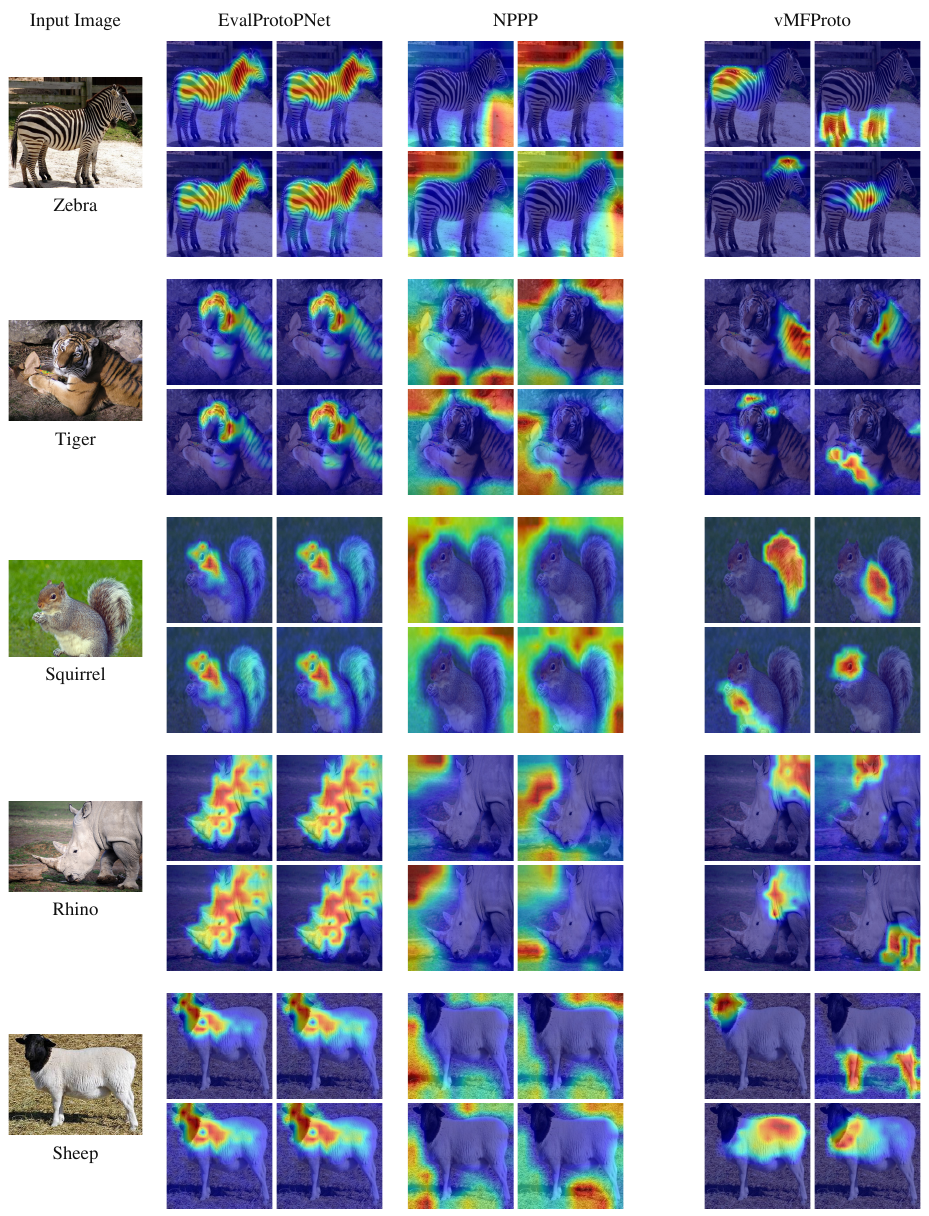}
  \caption{\textbf{Class-specific prototypes on AP-10K.} Each row shows the input image and the top four prototype activation maps for its ground-truth class. All methods use DINOv2 ViT-B/14 and $J{=}5$ prototypes per class.}
  \label{fig:supp_ap10k_qual}
\end{figure*}

\begin{figure*}[t]
  \centering
  \includegraphics[width=\textwidth,height=0.83\textheight,keepaspectratio]{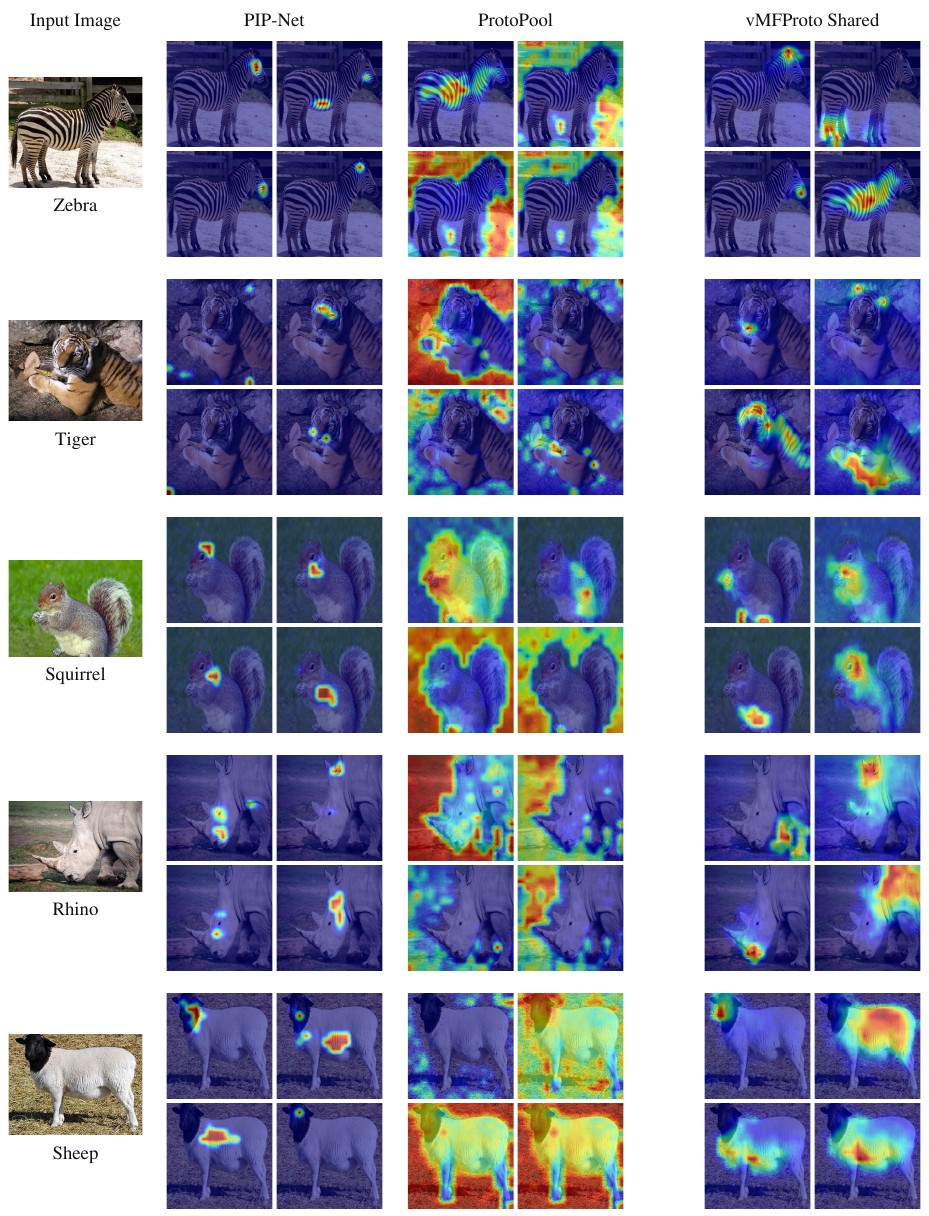}
  \caption{\textbf{Shared prototypes on AP-10K.} For the same images as in \cref{fig:supp_ap10k_qual}, we show each method's four highest-ranked shared prototypes for the ground-truth class. All methods use DINOv2 ViT-B/14 and a pool of $R{=}250$ prototypes.}
  \label{fig:supp_ap10k_shared_qual}
\end{figure*}

\begin{figure*}[t]
  \centering
  \includegraphics[width=\textwidth,height=0.83\textheight,keepaspectratio]{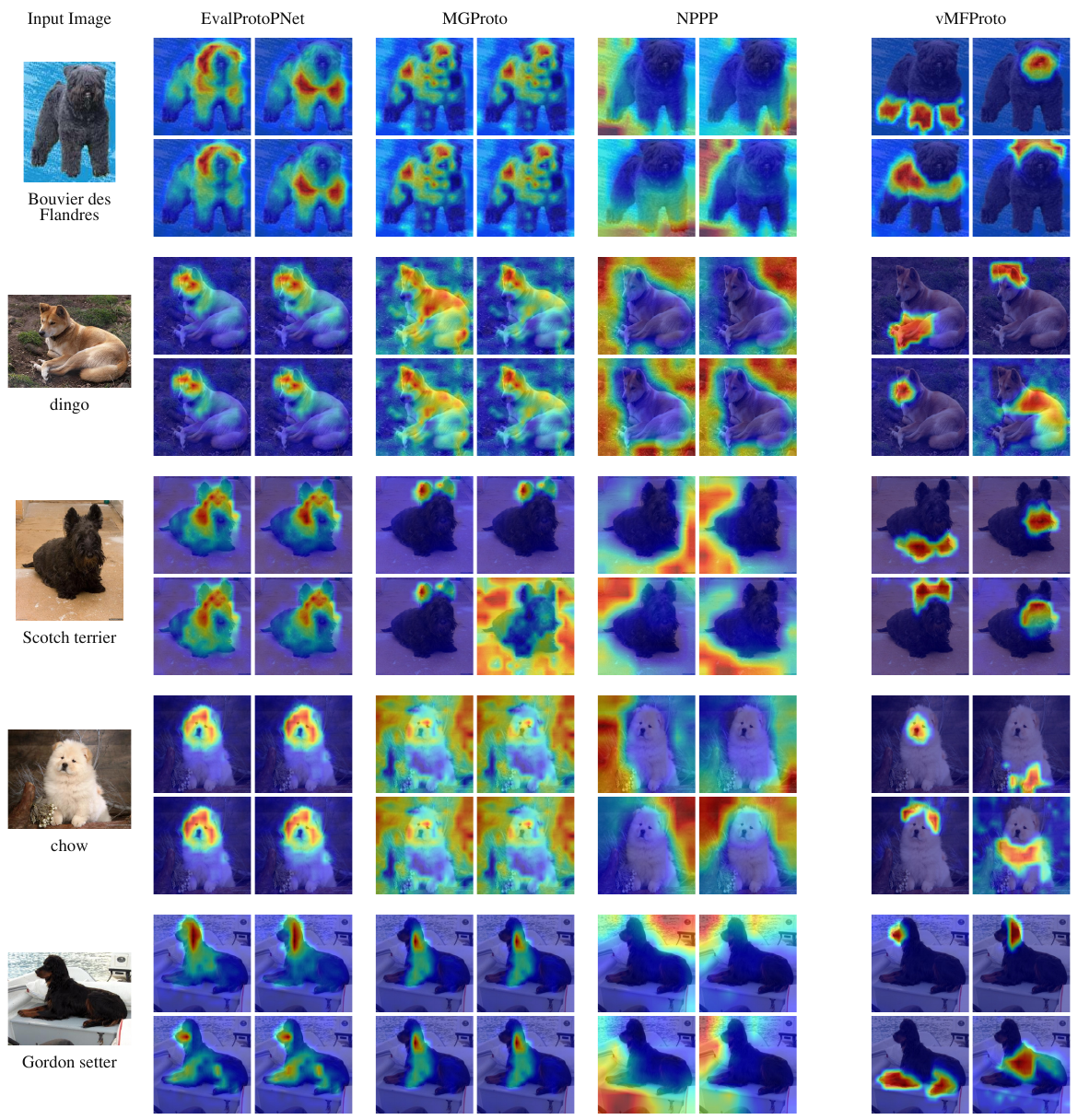}
  \caption{\textbf{Class-specific prototypes on Stanford Dogs.} Each row shows the input image and the top four prototype activation maps for its ground-truth class. All methods use DINOv2 ViT-B/14 and $J{=}5$ prototypes per class.}
  \label{fig:supp_dogs_qual}
\end{figure*}

\begin{figure*}[t]
  \centering
  \includegraphics[width=\textwidth,height=0.83\textheight,keepaspectratio]{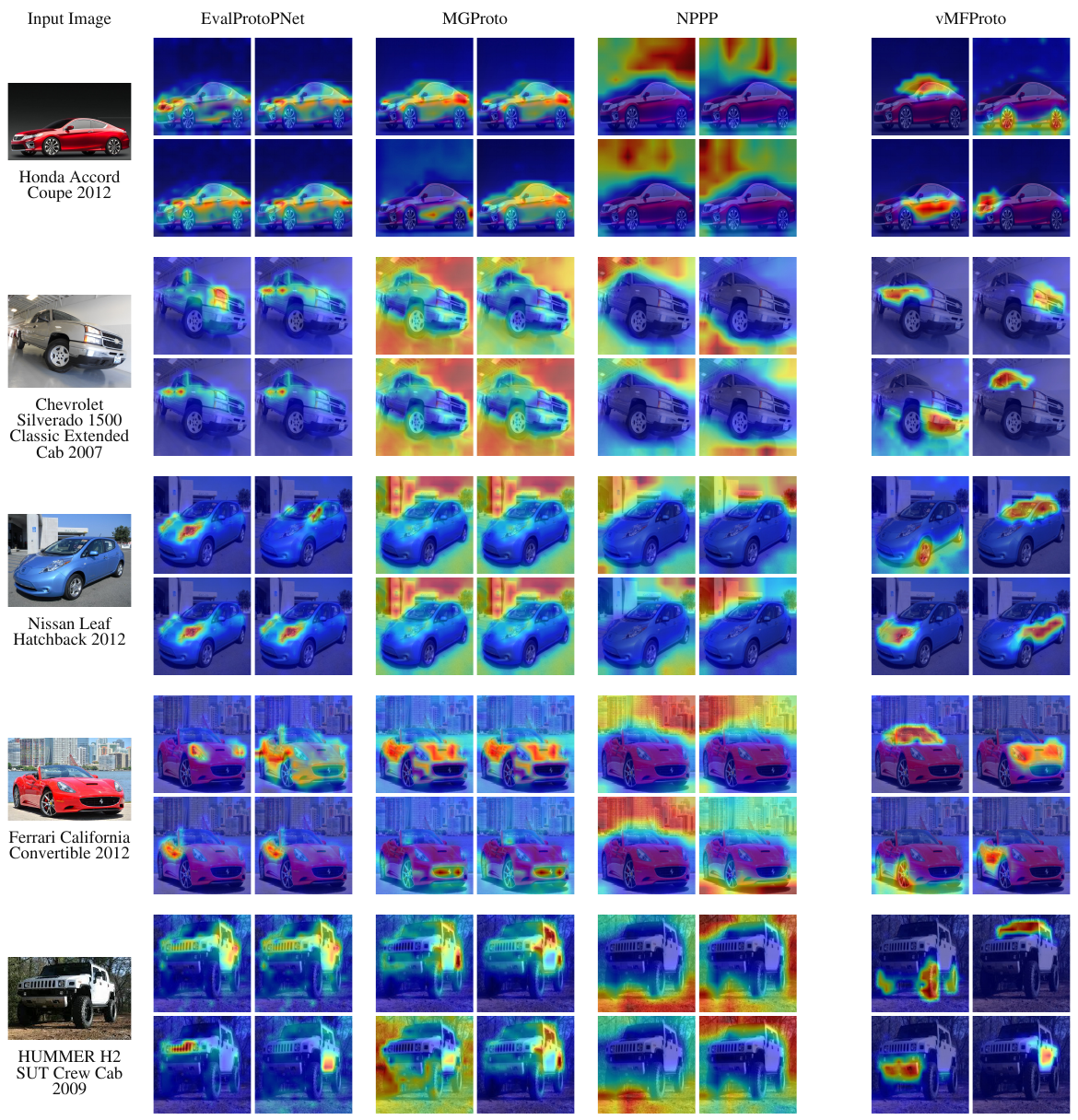}
  \caption{\textbf{Class-specific prototypes on Stanford Cars.} Each row shows the input image and the top four prototype activation maps for its ground-truth class. All methods use DINOv2 ViT-B/14 and $J{=}5$ prototypes per class.}
  \label{fig:supp_cars_qual}
\end{figure*}

Alongside the prediction-level why-tables, \cref{fig:supp_cub_qual,fig:supp_ap10k_qual,fig:supp_dogs_qual,fig:supp_cars_qual} compare class-specific models on five selected test images per dataset, while \cref{fig:supp_cub_shared_qual,fig:supp_ap10k_shared_qual} show shared-pool responses for the same CUB and AP-10K images. In each row, we overlay four prototype responses for the ground-truth class without a foreground mask, so that responses outside the object remain visible. In several examples, \method{} highlights separate regions where the baselines repeat a response. For instance, on the AP-10K squirrel, its class-specific prototypes highlight the head, forepaw, and tail, while EvalProtoPNet repeatedly responds around the head.

\FloatBarrier

\FloatBarrier

\end{document}